\documentclass{article}

\usepackage[preprint]{neurips_2026}

\usepackage[utf8]{inputenc}
\usepackage[T1]{fontenc}
\usepackage{url}
\usepackage{booktabs}
\usepackage{amsfonts}
\usepackage{amsmath, amssymb, amsthm}
\usepackage{nicefrac}
\usepackage{microtype}
\usepackage{graphicx}
\usepackage[dvipsnames]{xcolor}
\usepackage{float}
\usepackage{algorithm}
\usepackage{algorithmic}
\usepackage{wrapfig}

\newtheorem{theorem}{Theorem}
\newtheorem{corollary}[theorem]{Corollary}
\newtheorem{remark}{Remark}

\title{What Fits (Into Few Tokens) Doesn't Overfit:  Compression and Generalization in ML Research Agents}

\author{
  Martin Andres Bertran \\
  Amazon Responsible AI \\
  \And
  Aaron Roth \\
  Amazon Responsible AI \\
  University of Pennsylvania \\
  \And
  Zhiwei Steven Wu \\
  Amazon Responsible AI \\
  Carnegie Mellon University \\
}

\begin{document}

\maketitle

\begin{abstract}

Reusing a held-out benchmark adaptively should, in principle, invite overfitting. Yet benchmark-driven machine learning (ML) has produced surprisingly little overfitting in practice. An attractive hypothesis is that successful ML strategies are highly compressible. We study this in the setting of LLM-driven research agents, where the hypothesis becomes directly testable via two complementary information bottlenecks. In \emph{output compression}, an exploration agent adaptively searches for high-performance models using a validation set, and we test whether a fresh ``reproducer agent'' can reproduce its performance given only an extremely short prompt and the training data. In \emph{input compression}, the explorer receives only one-bit feedback indicating whether each submitted model improves on the running best. Across 8 datasets spanning tabular classification, vision, language modeling, diffusion modeling, and reward modeling, we find that these bottlenecks have little effect on performance: short prompts and compressible feedback are sufficient to reproduce and find high-performance models. The hypothesis is falsifiable: when we deliberately induce validation-set overfitting, the results fail to reproduce with short prompts. Taken together, our results support a description-length explanation for the lack of overfitting in benchmark-driven ML: successful strategies occupy a low-complexity region of strategy space.
\end{abstract}

\section{Introduction}

Held-out benchmarks are supposed to be used sparingly. In the textbook workflow, a model is trained and selected using training or validation data, and the final test set is touched only once. Machine learning practice looks very different: benchmarks are reused for years, with researchers iteratively fitting models, examining results, and choosing the next experiment in response. In principle, this adaptive reuse should invite overfitting: reported gains may reflect idiosyncrasies of the benchmark rather than real progress, the same pathology known in the empirical sciences as data dredging or p-hacking \citep{ioannidis2005most, gelman2016statistical, simmons2011false}.

Yet heavily reused ML benchmarks have often remained surprisingly informative. New-test-set studies and leaderboard meta-analyses find that improvements on old benchmarks largely transfer to fresh evaluations, despite extensive adaptive reuse \citep{recht2019imagenet,roelofs2019meta}. What keeps ordinary ML research from triggering this worst-case overfitting?

A natural explanation is that successful ML strategies are highly compressible. Although a researcher may observe many benchmark scores, the strategy that survives this process often consists of a short list of familiar choices: an architecture family, optimizer, schedule, data handling recipe, and regularization scheme. Following the description-length perspective of \citet{arora2021rip}, this shifts attention from the full adaptive transcript to the shortest description needed to communicate the successful strategy to a capable implementer with the usual background knowledge, but without the researcher's validation interactions. If that description is short, the final model's dependence on the benchmark is much smaller than the transcript of adaptive interactions suggests.

\begin{figure}[t!]
    \centering
    \includegraphics[width=\linewidth]{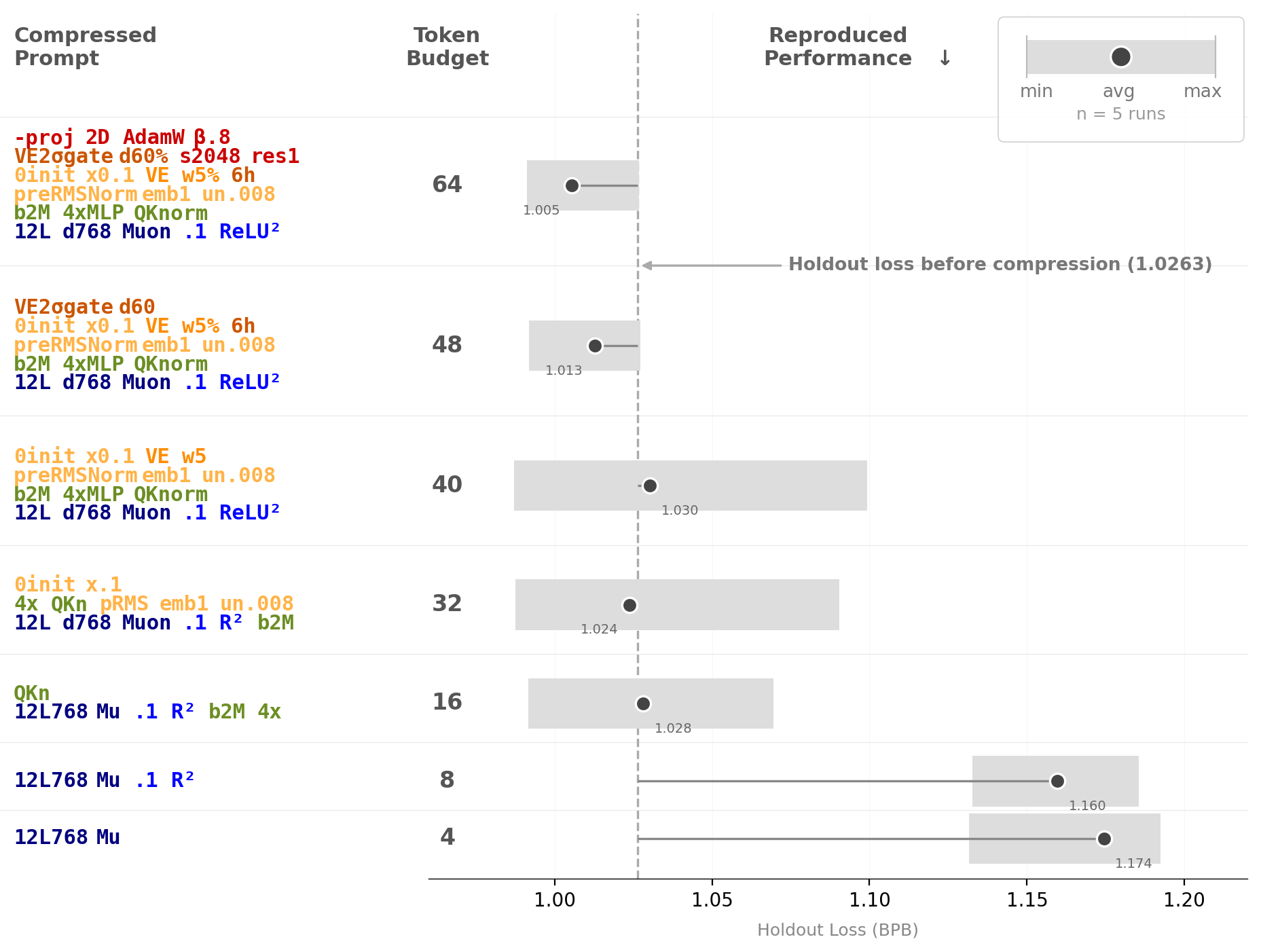}
    \caption{Compressibility of an autonomous agent's language model pre-training strategy. An explorer agent develops a 12-layer, 768-dimensional GPT with ${\sim}30$ non-default choices; we compress its strategy into progressively shorter prompts (64 down to 4 tokens) and hand each to 5 independent reproducers that implement the strategy without the explorer's code or validation set. \textbf{Left}: compressed prompts. Tokens encode architecture (\texttt{12L d768}), optimizer (\texttt{Muon .1}), activation (\texttt{ReLU\textsuperscript{2}}), batch size (\texttt{b2M}), and normalization (\texttt{QKnorm}); color indicates survival depth (dark blue = survives tightest budgets, red = dropped first). \textbf{Right}: reproducer holdout loss (BPB); dot = mean, gray band = min--max, dashed line = explorer before compression. Performance holds to 16 tokens; a cliff at 8 tokens coincides with loss of batch size, MLP ratio, and QK-norm.}
    \label{fig:hero}
\end{figure}

LLM-driven data science agents make this hypothesis directly testable. Unlike human researchers, agents can be reset: a fresh agent has no state from a previous run except what is passed in its prompt. We therefore place an explicit bottleneck between an \emph{explorer}, which adaptively searches using a validation set, and a \emph{reproducer}, which receives only a short prompt and the training data (no validation set, no explorer code, no transcript). If the reproducer nevertheless matches the explorer's performance, then the validation-dependent information needed to specify the strategy has fit through the prompt. We call this certificate \textbf{Output Compression}. Figure~\ref{fig:hero} illustrates the idea: a language-modeling strategy survives compression down to 16 tokens with no loss in holdout performance, enabled by the shared background knowledge that serves as an implicit prior.

%Figure~\ref{fig:hero} illustrates the idea: an explorer's language-modeling strategy, involving ${\sim}30$ non-default design choices, is compressed into progressively shorter prompts and handed to fresh reproducers. The strategy survives compression down to 16 tokens with no loss in holdout performance. Remarkably, LLM agents can communicate strategies without an explicit codebook; their shared background knowledge of ML concepts, architectures, and standard practices serves as an implicit prior that makes extreme compression possible. The formal guarantee applies to the reproducer's hypothesis; when the reproducer closely matches the explorer, the certificate is evidence that the explorer's strategy was itself compressible.

We also test compressibility from the other direction, by limiting what the explorer can observe about the validation set. In our \textbf{Input Compression} experiment, the explorer receives only one-bit feedback (a \emph{ladder mechanism}, following \citet{blum2015ladder}): whether each submitted model improves on the running best. The explorer may submit an adaptive sequence of models, but its validation-dependent view is just the resulting binary transcript. Thus the selected model is a function of the training data and a short transcript, not arbitrary numerical validation scores. Counting the possible transcripts gives a direct description-length bound and a simultaneous confidence interval over the ladder's improvement checkpoints.

\paragraph{Compression preserves performance and certifies progress.}
We evaluate Claude-based autonomous ML research agents on 8 datasets spanning tabular classification, image classification, language modeling, diffusion modeling, and reward modeling. Both bottlenecks preserve performance. Under output compression, 32-token prompts largely reproduce the explorer's adaptively optimized model performance; under input compression, one-bit ladder feedback matches or exceeds full numerical score feedback. On the classification tasks, the ladder mechanism also gives valid simultaneous confidence bounds on population accuracy for these adaptively optimized performance. Empirically, the intervals are tight enough to distinguish the best ladder-selected models from earlier checkpoints, certifying genuine progress rather than merely validation-set wins.

\paragraph{Compression fails under validation exploitation.}
The same agents can be steered into an overfitting regime by giving them validation-data access. When prompted to maximize validation performance at all costs, explorers reliably overfit: at 38 of 102 checkpoints, the validation metric was more than 10\% better than the corresponding holdout metric. The compression story predicts that such gains should not reproduce: if they come from idiosyncrasies of the validation set, a fresh reproducer without validation access should lose them. Consistent with that prediction, these validation-specific advantages disappear under prompt compression, which separates honest from exploiting checkpoints with 100\% sensitivity and 91\% specificity.

\paragraph{Related work.}
Our work builds on empirical studies showing that benchmark reuse often produces less overfitting than worst-case adaptive-overfitting theory predicts \citep{recht2019imagenet,roelofs2019meta}, and on structural explanations for this robustness based on restricted analyst behavior or restricted hypothesis classes \citep{mania2019model,feldman2019advantages,zrnic2019natural}. The formal backbone comes from adaptive data analysis, which shows that adaptively chosen hypotheses generalize when their dependence on reused data is mediated by a bounded information channel \citep{dwork2015preserving,dwork2015reusable,dwork2015generalization,blum2015ladder,bassily2016algorithmic,russo2016controlling,xu2017information,cummings2016adaptive,steinke2020reasoning,jung2020new}, and from compression-based generalization  \citep{arora2021rip, rissanen1978modeling,arora2018compression,akinwande2024understanding}. We bring these ideas to autonomous ML agents \citep{huang2024mlagentbench,chan2025mlebench,jiang2025aide,lu2024aiscientist}: in a resettable-agent setting, compression becomes both an empirical probe of why ordinary benchmark reuse generalizes and a falsifiable test for validation-set exploitation. Appendix~\ref{app:additional-related-work} gives a more detailed discussion.

\section{Preliminaries}\label{sec:prelim}

\subsection{Setting}

We consider an autonomous ML research agent $\mathcal{A}$, which we refer to as the \emph{explorer}, that iteratively develops a model for a machine learning task (prediction, language modeling, generative modeling, etc.). The explorer has access to two datasets drawn i.i.d.\ from a data distribution $\mathcal{P}$: a training set $D_{\mathrm{train}}$ of $N$ samples used for model training, and a validation set $D_{\mathrm{val}}$ of $n$ samples used for evaluating and comparing candidate models.

The explorer runs an iterative research loop: at each iteration $t = 1, \ldots, T$, it modifies a training script, trains a model $h_t$ on $D_{\mathrm{train}}$, and issues a \emph{query} asking for feedback on $h_t$ from $D_{\mathrm{val}}$. Based on the response (whose form we formalize in Section~\ref{sec:bottlenecks}), it decides what to try next. After $T$ queries, the explorer reports its best model $\hat{h}$ along with its validation risk $\hat{R}_{D_{\mathrm{val}}}(\hat{h})$. Let $R(h) = \mathbb{E}_{(x,y) \sim \mathcal{P}}[\ell(h(x), y)]$ denote the true risk (population loss) of a hypothesis $h$. After $T$ adaptive interactions with $D_{\mathrm{val}}$, the gap
\[
    R(\hat{h}) - \hat{R}_{D_{\mathrm{val}}}(\hat{h})
\]
can in principle be large: the explorer may achieve low validation loss by tuning, implicitly or explicitly, to the specific samples in $D_{\mathrm{val}}$ rather than by finding a model that generalizes. Yet heavily reused ML benchmarks often remain informative in practice. We ask what property of natural ML research keeps this gap small, and how to test it directly.

\subsection{Two Information Bottlenecks}\label{sec:bottlenecks}

We probe the compressibility of the explorer's strategy by narrowing the $D_{\mathrm{val}}$ channel on either side: compressing what the explorer \emph{outputs} (its final strategy) or what it \emph{sees} (feedback from $D_{\mathrm{val}}$). In the output-bottleneck experiment, only a short message about the explorer's strategy is passed to a fresh reproducer. In the input-bottleneck experiment, the explorer itself receives only a short binary transcript of validation feedback. In both cases, the final model depends on $D_{\mathrm{val}}$ only through a deliberately low-information channel.

\paragraph{Output compression: prompt-and-reproduce.}
Given any hypothesis $\hat{h}$ produced by the explorer, a separate compressor reads the exploration transcript and distills the strategy behind $\hat{h}$ into a short message $m$. A fresh agent $\mathcal{B}$ (the \emph{reproducer}) receives only $m$ together with the training set $D_{\mathrm{train}}$; it has no access to $D_{\mathrm{val}}$, the explorer's code, or the transcript. The reproducer implements the strategy in its own training script and produces a hypothesis $h_m = \mathcal{B}(m, D_{\mathrm{train}})$, so the message $m$ is the only route by which information from $D_{\mathrm{val}}$ can reach the reproducer's model.

In experiments, $m$ is a short prompt written by an LM compressor and decoded by an LM reproducer. This works because the LM prior is powerful: a few tokens can invoke shared knowledge about architectures, optimizers, schedules, and training conventions.

\paragraph{Input compression: the ladder mechanism.}
\begin{wrapfigure}{r}{0.5\columnwidth}
\vspace{-1.2em}
\begin{minipage}{0.5\columnwidth}
\begin{algorithm}[H]
\caption{Ladder Mechanism (one-bit variant; cf.\ \citealt{blum2015ladder})}\label{def:ladder}
\small
\begin{algorithmic}[1]
\REQUIRE Query budget $T_{\max}$, improvement budget $K_{\max}$, validation set $D_{\mathrm{val}}$, empirical risk $\hat R_{D_{\mathrm{val}}}$ to minimize
\STATE $h^\star \gets \texttt{null}$, $r^\star \gets +\infty$, $K \gets 0$, $\sigma \gets \texttt{[]}$
\FOR{$t = 1, \ldots, T_{\max}$}
    \STATE Explorer proposes $h_t$ (adaptive in $\sigma$, $D_{\mathrm{train}}$)
    \STATE $r_t \gets \hat{R}_{D_{\mathrm{val}}}(h_t)$
    \IF{$r_t < r^\star$}
        \STATE $h^\star \!\gets\! h_t$, $r^\star \!\gets\! r_t$, $K \!\gets\! K\!+\!1$, append $1$ to $\sigma$
    \ELSE
        \STATE Append $0$ to $\sigma$
    \ENDIF
    \STATE Return $\sigma_t$ to explorer
    \IF{$K = K_{\max}$}
        \STATE \textbf{break}
    \ENDIF
\ENDFOR
\RETURN $h^\star$, $\sigma$
\end{algorithmic}
\end{algorithm}
\end{minipage}
\vspace{-1em}
\end{wrapfigure}
A complementary bottleneck restricts what each query can reveal. Instead of returning a numerical score, we answer each of the explorer's queries with a single bit: ``did this model improve on the running best, or not?'' The explorer still searches adaptively, but its entire view of $D_{\mathrm{val}}$ over a run is a binary string of ``improved'' / ``not improved'' responses. We call this the \emph{ladder mechanism}, by analogy with the Ladder of \citet{blum2015ladder}. Both mechanisms share a single design idea (the running best updates only when a new model beats it), but differ in what they return: the mechanism of \cite{blum2015ladder} releases a rounded numerical score, whereas ours returns a single bit. We establish the generalization guarantee in Section~\ref{sec:ladder}. Thus the explorer's view of $D_{\mathrm{val}}$ is the binary transcript $\sigma \in \{0,1\}^T$; its final hypothesis $\hat{h}$ is a function of $(\sigma, D_{\mathrm{train}})$ alone, and depends on $D_{\mathrm{val}}$ only through the bits of $\sigma$.

\section{Compression and Generalization}\label{sec:theory}

Our explanation for why the validation-population risk gap stays small rests on a classical principle: compressible hypotheses generalize. Concretely, when a procedure's output depends on $D_{\mathrm{val}}$ only through a member of a finite description set $\mathcal{M}$ fixed in advance, $\log_2 |\mathcal{M}|$ is an upper bound on the description length of the resulting hypothesis; at most $|\mathcal{M}|$ hypotheses could ever arise, and Hoeffding plus a union bound gives the standard rate.

Throughout this section, all statements condition on $D_{\mathrm{train}}$ and on any randomness independent of $D_{\mathrm{val}}$ (e.g., training seeds, reproducer randomness, etc.); after conditioning, decoders are fixed functions of the message or transcript. Since the conditional high-probability statements hold for every fixed value of this independent randomness, they also hold after integrating over it.

% Throughout this section, statements are conditional on $D_{\mathrm{train}}$ and on any randomness drawn independently of $D_{\mathrm{val}}$ (for example, training seeds, agent sampling randomness, or reproducer randomness). After conditioning, the relevant decoders are fixed functions of the compressed message or transcript. Since the conditional high-probability statements hold for every fixed value of this independent randomness, they also hold after integrating over it.

\begin{theorem}[Adaptive Description-Length Generalization; \citealt{dwork2015preserving,dwork2015reusable}]\label{thm:meta}
Let $\mathcal{M}$ be a finite set of descriptions, fixed before observing $D_{\mathrm{val}}$, and let $\Phi : \mathcal{M} \to \mathcal{H}$ be a fixed decoder, so each $m \in \mathcal{M}$ determines a hypothesis $h_m := \Phi(m)$ without reference to $D_{\mathrm{val}}$. Assume $\ell \in [0,1]$. Then for any $\delta > 0$, with probability at least $1 - \delta$ over $D_{\mathrm{val}}$,
\[
    \sup_{m \in \mathcal{M}} \big| \hat{R}_{D_{\mathrm{val}}}(h_m) - R(h_m) \big| \;\leq\; \sqrt{\tfrac{\log(2|\mathcal{M}|/\delta)}{2n}}.
\]
\end{theorem}

The two bottlenecks of Section~\ref{sec:bottlenecks} are instantiations of Theorem~\ref{thm:meta}, differing only in their choice of $(\mathcal{M}, \Phi)$: the compressed prompt given to the reproducer (Section~\ref{sec:theory:output}) and the binary transcript produced by the ladder mechanism (Section~\ref{sec:ladder}). The empirical question of whether natural research strategies actually fit inside short descriptions then becomes testable in Section~\ref{sec:experiments}.

\subsection{Output Compression}\label{sec:theory:output}

In the output bottleneck, the only validation-dependent object passed to the reproducer is the message $m$. Once $m$ is fixed, the reproducer's hypothesis $h_m = \mathcal{B}(m, D_{\mathrm{train}})$ depends only on $m$, $D_{\mathrm{train}}$, and randomness independent of $D_{\mathrm{val}}$. Thus a token budget induces a finite message class: if $m$ has at most $B$ tokens from vocabulary $V$, then $\mathcal{M}_B := V^{\leq B}$ and $|\mathcal{M}_B| \leq |V|^{B+1}$. Applying Theorem~\ref{thm:meta} gives:

\begin{corollary}[Output Compression Bound]\label{thm:union}
Fix token budget $B$ and vocabulary $V$ before observing $D_{\mathrm{val}}$, and let $\mathcal{A}$ be any (possibly adaptive) procedure that interacts with $D_{\mathrm{val}}$ and outputs a message $m$ of at most $B$ tokens in $V$. Let $h_m = \mathcal{B}(m, D_{\mathrm{train}})$ for a decoder $\mathcal{B}$ fixed under the conditioning convention above. Under the assumptions of Theorem~\ref{thm:meta}, with probability at least $1 - \delta$ over $D_{\mathrm{val}}$,
\[
    \big| \hat{R}_{D_{\mathrm{val}}}(h_m) - R(h_m) \big| \;\leq\; \sqrt{\frac{(B+1)\ln|V| + \ln(2/\delta)}{2n}}.
\]
\end{corollary}

This bound constrains the \emph{reproducer's} hypothesis $h_m$, whose only validation-dependent input is the short message $m$; it does not directly constrain the explorer's original hypothesis $\hat h$. If $h_m$ nevertheless matches the explorer's validation performance, this is evidence that the explorer's strategy can be distilled into a short description whose decoded model preserves performance and satisfies the generalization bound.

This uniform bound is loose because it assigns equal mass to every $B$-token string, while the compressor writes in a dense ML shorthand (Figure~\ref{fig:hero}) drawn from a much smaller set of choices. We therefore treat $B$ as an operational proxy for description length: it is the quantity the experiment directly controls, and varying it probes the degree of compression. A sharper PAC-Bayes analysis could replace $(B+1)\ln|V|$ with $-\ln p_\text{LM}(m)$ under an appropriate language-model prior over messages, following \citet{akinwande2024understanding}; how to construct such a prior for agent-written strategy prompts remains an open question.

% BACKUP: previous version of the paragraph above, kept in case we want to revert.
% The $B$ in Corollary~\ref{thm:union} is literally the token budget used in experiments; no codebook is designed or enumerated. The bound is loose in absolute terms, however, because it assigns equal mass to every $B$-token string, whereas almost all such strings are incoherent and would be heavily downweighted by any language model. A PAC-Bayes version using a language-model prior over messages would replace $(B+1)\ln|V|$ with $-\ln p_\text{LM}(m)$, giving the kind of non-vacuous bound developed by \citet{akinwande2024understanding} for static prompt engineering; we leave a quantitative instantiation to future work.

\subsection{The Ladder Mechanism}\label{sec:ladder}

The ladder mechanism bottlenecks the explorer's \emph{input}: after $T_{\max}$ queries and at most $K_{\max}$ improvements, the explorer's validation-dependent view is only a binary transcript. The amount of information revealed by the $j$-th improvement checkpoint is not the full $T_{\max}$ bits, but the locations of the first $j$ improvements. There are at most
\[
    N_j := \binom{T_{\max} - 1}{\,j-1\,}
\]
such transcripts, since the first model necessarily improves on the null running best. A union bound over these transcripts gives a simultaneous confidence interval for every improvement checkpoint.

\begin{corollary}[Ladder Generalization Bound]\label{thm:ladder}
Fix budgets $T_{\max}$ and $K_{\max}$, and confidence levels $(\delta_1, \ldots, \delta_{K_{\max}})$ with $\sum_j \delta_j \leq \delta$, before observing $D_{\mathrm{val}}$. Run the one-bit ladder mechanism for at most $T_{\max}$ queries and stop after at most $K_{\max}$ improvements. Let $\hat{h}^{(j)}$ denote the running-best model at the $j$-th improvement checkpoint, for each realized $j \leq K$, and define $N_j = \binom{T_{\max}-1}{j-1}$. Assume $\ell \in [0,1]$. Then with probability at least $1 - \delta$ over $D_{\mathrm{val}}$, simultaneously for every realized checkpoint $j \in \{1, \ldots, K\}$,
\[
    \big| \hat{R}_{D_{\mathrm{val}}}(\hat{h}^{(j)}) - R(\hat{h}^{(j)}) \big| \;\leq\; \varepsilon_j \;:=\; \sqrt{\frac{\ln(2 N_j / \delta_j)}{2n}} \;=\; O\!\left(\sqrt{\frac{j \log T_{\max} + \log(1/\delta_j)}{n}}\right)
\]
\end{corollary}

Unlike Corollary~\ref{thm:union}, which constrains the reproducer's hypothesis, Corollary~\ref{thm:ladder} constrains the explorer's own running-best model at every improvement checkpoint (proof in Appendix~\ref{app:ladder_proof}). Because $N_j$ grows with $j$, early checkpoints get tighter bounds: under our experimental parameters ($K_{\max} = 7$, $T_{\max} = 50$, $n = 5{,}000$, $\delta = 0.05$), the first checkpoint has $N_1 = 1$ and width $\varepsilon_1 \approx 2.4$pp while the last has $N_7 = \binom{49}{6} \approx 14\times 10^6$ and $\varepsilon_7 \approx 4.7$pp.

% Unlike Corollary~\ref{thm:union}, which constrains the reproducer's hypothesis, Corollary~\ref{thm:ladder} constrains the explorer's own running-best model at every improvement checkpoint. Because the stratum sizes grow with $j$, early checkpoints get tighter bounds: at $K_{\max} = 7$, $T_{\max} = 50$, $n = 5{,}000$, $\delta = 0.05$, and uniform allocation $\delta_j = \delta/K_{\max}$, the first checkpoint has $N_1 = 1$ and width $\varepsilon_1 \approx 2.4$pp while the last has $N_7 = \binom{49}{6} \approx 14\times 10^6$ and $\varepsilon_7 \approx 4.7$pp.

For classification (Bernoulli loss) the Chernoff-optimized binomial-KL tail replaces Hoeffding pointwise in the proof and tightens $\varepsilon_j$ by up to $\approx 2\times$ when the explorer's accuracy is bounded away from $1/2$; we use this tighter form for the Figure~\ref{fig:ladder_bounds} CIs and develop it in Appendix~\ref{app:chernoff}.

\section{Experiments}\label{sec:experiments}

\paragraph{Datasets.} We evaluate on 8 tasks chosen to span the range of modern ML: tabular classification, vision, language modeling, a generative diffusion task, and reward modeling. Table~\ref{tab:datasets} summarizes the setup. For each task we construct disjoint training, validation, and holdout splits; only the validation split is exposed through the permitted feedback channel, and the holdout is used only for post hoc evaluation by the experiment harness. Appendix~\ref{app:datasets} gives the dataset-specific split construction.

%For the five classification datasets the per-example validation and holdout samples are approximately i.i.d.\ from the evaluation distribution, which is the regime our formal bounds require; the three other datasets deviate from this (WikiText-103 BPB is a token-level average with within-document correlation, and the CIFAR-100 diffusion and HH-RLHF reward-model splits come from pre-existing task-specific evaluation pools) and we treat those empirically. 

% \paragraph{Scope of the formal guarantee.} Corollaries~\ref{thm:union} and \ref{thm:ladder} assume a bounded loss $\ell \in [0,1]$ and an i.i.d.\ sample of $n$ evaluation units. Both assumptions hold naturally on the five classification datasets, where the loss is the $0/1$ error and the evaluation unit is one validation example. The three other tasks do not satisfy these assumptions without modification: reward-model log-loss is unbounded without clipping; WikiText-103 BPB is a token-level average rather than a per-example statistic; and CIFAR-100 Diffusion MSE requires a declared range. We therefore present the val--holdout behavior on those three datasets as empirical evidence of the compression story, and restrict the formal generalization bounds to the five classification datasets.

\begin{table}[h!]
\centering
\caption{Datasets. Holdout is used only for post hoc evaluation. Metrics with ``$\uparrow$'' are higher-is-better (accuracy); ``$\downarrow$'' are lower-is-better (log-loss, MSE, BPB). Full dataset provenance, licensing, and split construction are in Appendix~\ref{app:datasets}.}
\label{tab:datasets}
\small
\begin{tabular}{lllrrr}
\toprule
Dataset & Domain & Metric & $n_{\mathrm{train}}$ & $n_{\mathrm{val}}$ & $n_{\mathrm{holdout}}$ \\
\midrule
Folktables (ACSIncome) & Tabular (census)    & Accuracy $\uparrow$   & 150K  & 50{,}000 & 50{,}000 \\
Gene-Expr              & High-dim tabular    & Accuracy $\uparrow$   & 20K   & 20{,}000 & 20{,}000 \\
SST-2                  & NLP sentiment       & Accuracy $\uparrow$   & 67K   & 20{,}000 & 20{,}000 \\
CIFAR-10            & Vision classification      & Accuracy $\uparrow$   & 50K   & 5{,}000  & 5{,}000  \\
CIFAR-100 Diffusion    & Generative          & MSE $\downarrow$      & 50K   & 5{,}000  & 5{,}000  \\
ImageNet-1K            & Vision (1000-class) & Accuracy $\uparrow$   & 1.28M & 25{,}000 & 25{,}000 \\
Reward model (HH-RLHF) & Reward modeling     & Log-loss $\downarrow$ & 160K  & 5{,}000  & 5{,}000  \\
WikiText-103           & Language modeling   & BPB $\downarrow$      & 100M tok & 5M tok & 5M tok \\
\bottomrule
\end{tabular}
\end{table}

% \paragraph{Agent and pipeline.} Every agent role (explorer, compressor, reproducer) is a single Claude Opus model instance driven by Claude Code. The explorer receives a natural-language problem description, starter code for a training script, and access to the training and validation splits. It iteratively edits the training script, trains a model on $D_{\mathrm{train}}$, evaluates on $D_{\mathrm{val}}$, and decides what to change next. We record each \emph{improvement checkpoint} (an iteration that produces a new best validation metric) for downstream evaluation. The compressor is given the full explorer transcript and writes a natural-language prompt of at most $B$ tokens ($B \in \{32, 64\}$) describing the strategy. The reproducer receives only the compressed prompt and $D_{\mathrm{train}}$; it has no access to the explorer's code, no access to the transcript, and no access to $D_{\mathrm{val}}$. It writes a training script from scratch and produces a single model $h_m$ whose validation risk $\hat{R}_{D_{\mathrm{val}}}(h_m)$ we compare against the explorer's $\hat{R}_{D_{\mathrm{val}}}(\hat{h})$.

\paragraph{Agent and pipeline.} Every agent role (explorer, compressor, reproducer) is a single Claude Opus model instance driven by Claude Code. The explorer receives a problem description and starter training code, then iteratively edits the code, trains models on $D_{\mathrm{train}}$, and queries validation feedback. We record each \emph{improvement checkpoint} (a new best validation metric) for downstream evaluation. The compressor reads the explorer transcript and writes a short prompt of at most $B$ tokens; the main output-compression sweep uses $B \in \{32, 64\}$. Only the final token-limited prompt is passed to the reproducer. The reproducer receives this prompt and $D_{\mathrm{train}}$, modifies starter training code in an isolated workspace, and produces a model $h_m$ whose validation risk is compared against the explorer's. Full role prompts and compressor audit/dry-run procedures are in Appendix~\ref{app:prompt-templates}.

\paragraph{Access control.} The bottlenecks are enforced by the experiment harness rather than by prompt instructions alone. Validation examples are not present in the agent-visible workspace; agents access $D_{\mathrm{val}}$ only through an evaluation entry point, which returns either a scalar metric or the ladder's one-bit improvement signal. Reproducer agents run in a separate workspace containing $D_{\mathrm{train}}$ and the compressed prompt, but no explorer code, transcript, validation files, or validation API.

\paragraph{Framing.} Unless stated otherwise we use a neutral prompt: ``Try a range of approaches. Let the metric decide.'' This avoids nudging the explorer toward aggressive validation exploitation; Section~\ref{sec:detection} revisits the effect of prompting under unrestricted validation access.

\paragraph{Experimental conditions.} We use scalar validation feedback for the output-compression experiment: the explorer sees only the numerical metric for each submitted model, and its improvement checkpoints are later compressed into prompts for reproduction. For the input-compression experiment, we replace scalar feedback with the one-bit ladder mechanism and compare the resulting search trajectory to the scalar-feedback runs.

\subsection{Output Compression: Prompt Bottleneck on the Explorer's Strategy}\label{sec:exp:output}

\begin{figure}[t]
    \centering
    \includegraphics[width=\textwidth]{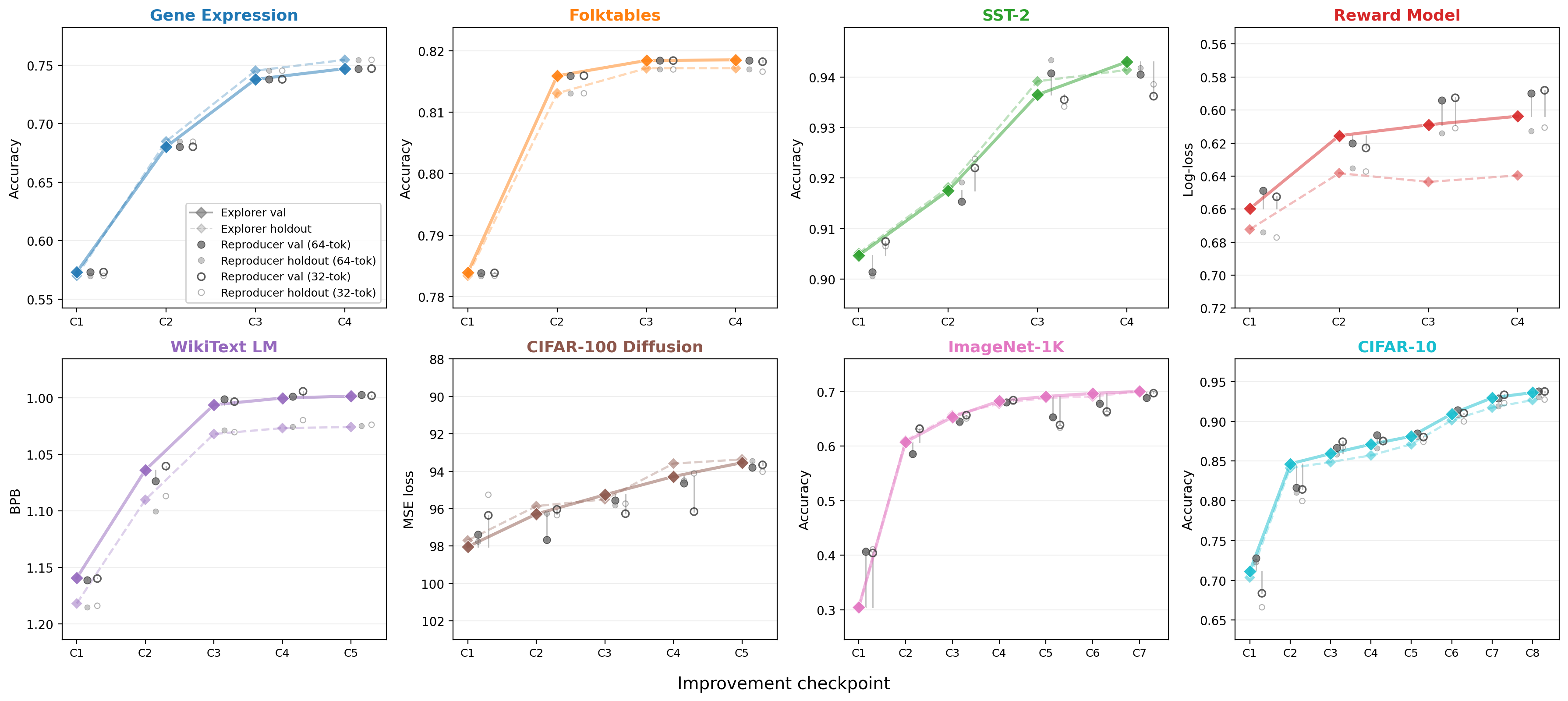}
    % \caption{Explorer trajectory with compressed reproducers across all 8 datasets. Each panel plots the explorer's progress at its \emph{improvement checkpoints}: the iterations at which the explorer achieved a new best validation metric during its run. Solid diamonds show the explorer's validation metric at each such checkpoint; faded diamonds show the corresponding holdout metric (not observed by the explorer). Filled and open circles show the reproducer's validation and holdout metrics at 64-token and 32-token compressed prompts, respectively. Connecting lines highlight the compression gap at each checkpoint. Across all datasets, reproducers closely track the explorer: the strategies discovered under score-based feedback compress into short prompts with minimal information loss.}
    \caption{Explorer trajectory with compressed reproducers across all 8 datasets. Each panel plots the explorer's improvement checkpoints: iterations achieving a new best validation metric. Solid diamonds: explorer validation; faded diamonds: explorer holdout (not observed by the explorer). Filled and open circles: reproducer validation at 64-token and 32-token prompts. Across all datasets, reproducers closely track the explorer: the strategies discovered under score-based feedback compress into short prompts with minimal information loss.}
    \label{fig:trajectory}

\end{figure}

We first bottleneck the \emph{output} of the explorer. Under score-based feedback, the explorer runs for up to several hundred iterations on each task, and every improvement checkpoint is passed to the compressor at both $B = 64$ and $B = 32$ tokens. Each compressed prompt is handed to a fresh reproducer with no validation access. Figure~\ref{fig:trajectory} plots, for each dataset, the explorer's validation and holdout metrics together with the reproducer's validation metric at every improvement checkpoint.

Figure~\ref{fig:trajectory} shows two patterns. First, the explorer's validation and holdout curves track closely throughout training on all 8 datasets; no overfitting emerges under score-based feedback alone. Second, reproducer validation metrics sit on top of, or slightly below, the explorer's validation metrics at nearly every checkpoint, including the very first ones where the strategy is still evolving. Across 41 improvement checkpoints, 32-token reproducers match or exceed the explorer under our one-sided 5\% relative-gap criterion on 38 checkpoints (92.7\%); the remaining failures occur on ImageNet and CIFAR-10 checkpoints where the short prompt misses enough detail to match the explorer under the threshold. Appendix~\ref{app:numeric-tables} gives the per-dataset counts for both token budgets. Appendix~\ref{app:example_prompts} shows representative 64-token prompts to give a sense of what the compressor actually produces. An empty-prompt baseline confirms that this information transfer is genuine: without a strategy description, the reproducer recovers only first-iteration performance (Appendix~\ref{app:empty-prompt}).

% The single 128-token failure is on CIFAR-10, where a complex schedule with many precisely tuned parameters exceeds the token ceiling, which is an incompressibility-by-genuine-complexity case of the kind discussed after Corollary~\ref{thm:union}; the additional 64-token failure is on WikiText-103's most complex checkpoint, which involves a custom optimizer with dozens of coupled hyperparameters. Aside from these two complexity-ceiling cases, the descriptions the explorer actually uses fit comfortably inside 64 tokens.

\subsection{Input Compression: Binary Feedback via the Ladder}\label{sec:exp:input}

% The output-compression result provides evidence that the explorer's final strategy is compressible. A complementary question is whether the explorer actually \emph{uses} the numerical precision of the score-based feedback during search, or whether the sign of improvement alone would have been enough. We answer this by rerunning each task under the ladder mechanism (Section~\ref{sec:ladder}). All other aspects (starter code, neutral framing, validation/holdout splits, compute budget) match the score-based condition.

Output compression shows the explorer's final strategy is compressible; we now ask whether it actually \emph{uses} the numerical precision of score-based feedback, or whether the sign of improvement alone suffices. We test this by rerunning each task under the ladder mechanism (Section~\ref{sec:ladder}), with all other aspects matching the score-based condition.

% The ladder explorer matches or slightly exceeds the score-based explorer on holdout performance on all 8 datasets, despite receiving $\approx 32\times$ less information per query; per-dataset trajectories for both conditions (three runs each) are in Appendix~\ref{app:ladder_trajectories}. The implication is that the search the agent performs is \emph{coarse}: it is choosing among a small number of qualitatively different strategies (architecture, optimizer, schedule), each of which is either clearly better or clearly worse than the running best. Numerical precision beyond a single bit per query does not translate into additional useful search. Combined with the output-compression evidence, this paints a coherent picture: the information that flows between $D_{\mathrm{val}}$ and the agent's final strategy is genuinely narrow in both directions.

The ladder explorer matches or slightly exceeds the score-based explorer on holdout performance on all 8 datasets, despite replacing numerical validation scores with a single bit of feedback per query; per-dataset trajectories for both conditions (three runs each) are in Appendix~\ref{app:ladder_trajectories}. In these runs, one-bit improvement feedback was sufficient to guide the agent toward the same kinds of high-performing strategies found under scalar feedback. This suggests that much of the useful validation signal is coarse: the agent often chooses among qualitatively different strategies whose effects are large enough to be detected by improvement/no-improvement feedback alone. Together with the output-compression evidence, this suggests that the information flowing between $D_{\mathrm{val}}$ and the agent's final strategy is narrow in both directions.

Finally, the ladder admits a direct generalization bound on the tasks that satisfy Corollary~\ref{thm:ladder}. With a pre-specified budget of $K_{\max} = 7$ improvements and $T_{\max} = 50$ queries, we obtain 95\% confidence intervals whose half-width ranges from $\approx 0.6$ to $\approx 2.5$pp depending on the dataset and checkpoint. The observed validation--holdout gaps (0.0--1.9pp) sit well inside these bounds in all 5 cases, with holdout serving as an independent empirical proxy for population performance rather than as the quantity covered by the theorem (Figure~\ref{fig:ladder_bounds}; numeric summary in Appendix~\ref{app:numeric-tables}). These intervals use the Chernoff-optimized form (Appendix~\ref{app:chernoff}), which tightens Corollary~\ref{thm:ladder} by up to $\approx 2\times$ at high accuracies. Because the bound holds simultaneously across all checkpoints, non-overlapping CIs certify that genuine progress occurred.

\begin{figure}[h!]
    \centering
    \includegraphics[width=\textwidth]{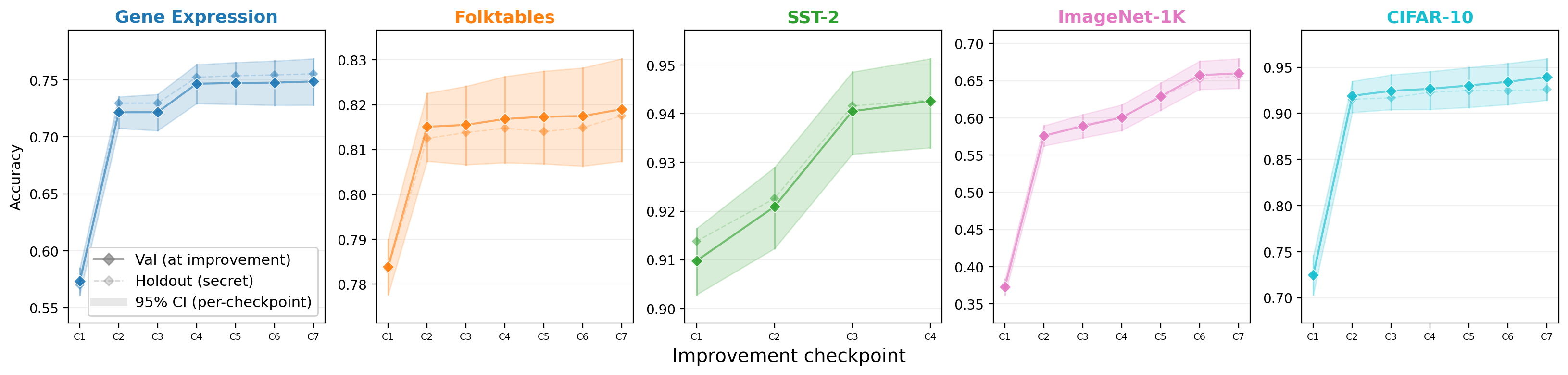}
    \caption{Ladder generalization bounds on the 5 classification datasets. At each improvement checkpoint, the solid line shows the explorer's validation metric and the dashed line shows the holdout metric (not observed by the explorer). Shaded bands show the 95\% confidence interval from the Chernoff-optimized (Bernoulli KL) form of Corollary~\ref{thm:ladder}, with $K_{\max} = 7$, $T_{\max} = 50$, and $\delta_j = \delta/K_{\max}$ (see Appendix~\ref{app:chernoff}). The corollary controls population performance; the holdout curve is shown as an independent empirical check.}
    \label{fig:ladder_bounds}
\end{figure}

\section{A Falsifiable Prediction: Validation-Specific Advantages Don't Reproduce}\label{sec:detection}

The compressibility story makes a falsifiable prediction: if an explorer's advantage comes from exploiting specifics of $D_{\mathrm{val}}$ rather than a genuinely good strategy, a reproducer working from a short prompt and $D_{\mathrm{train}}$ alone should not be able to reproduce that advantage. The $D_{\mathrm{val}}$-specific information cannot fit in a handful of tokens, and even if it did, the reproducer lacks $D_{\mathrm{val}}$ to act on it. Whatever the compressed prompt transmits, only components that operate on the training data produce value on the other side.

We stress-test this prediction in an intentionally strong exploitation regime: aggressive prompting (``maximize validation accuracy at all costs''), full sample-level validation access, and for several tasks smaller validation sets or injected label noise that reward memorization (Appendix~\ref{app:overfitting-details}). The goal is not to model every subtle form of benchmark overuse, but to test whether compression fails when validation-specific information is easy to exploit. Under these conditions, agents reliably exploit $D_{\mathrm{val}}$ at training time (Appendix~\ref{app:overfitting-details} gives per-task mechanisms); across 102 checkpoints, 38 exceed a 10\% relative validation--holdout gap.

% \paragraph{Result.} Calling a checkpoint \emph{overfitting} when the val--holdout gap exceeds 10\% relative, and the certificate \emph{failed} when the val--reproducer gap exceeds 5\% relative (both thresholds pre-specified), 128-token prompt-and-reproduce flags every overfitting checkpoint (38/38 sensitivity) and accepts 91\% of honest ones (58/64 specificity). The 6 false positives are strategies with high genuine description complexity (e.g., custom language-model training recipes), not validation exploitation --- the failure mode anticipated after Corollary~\ref{thm:union}. Increasing the budget to 128 tokens does not close the gap at exploiting checkpoints: what's missing is access to $D_{\mathrm{val}}$ at training time, not prompt capacity. Figure~\ref{fig:full_access_trajectory} shows per-checkpoint trajectories; Appendix~\ref{app:overfitting-details} gives the full detection rule and threshold sensitivity.

\begin{figure}[h!]
    \centering
    \includegraphics[width=\textwidth]{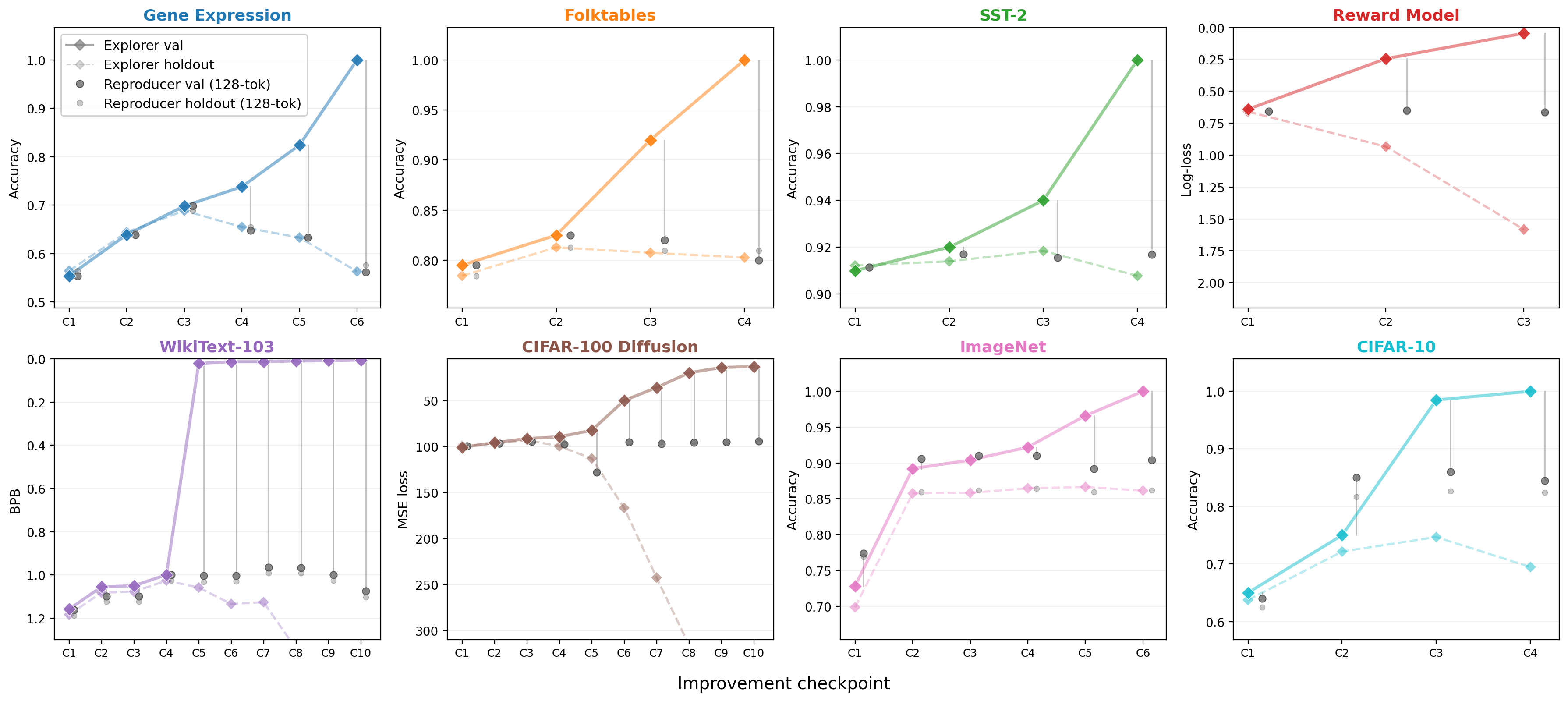}
    \caption{Validation-specific gains fail to reproduce under compression. Under aggressive prompting and sample-level validation access, explorers often continue improving validation performance after holdout performance stalls or degrades. Fresh reproducers given only a 128-token strategy prompt track the holdout behavior better than the inflated validation metric; the reproducer--explorer gap appears at the same checkpoints where validation and holdout diverge. Solid: explorer validation; faded diamonds: explorer holdout; circles: reproducer validation and holdout.}
    \label{fig:full_access_trajectory}
\end{figure}

 % \paragraph{Result.} 
Figure~\ref{fig:full_access_trajectory} shows the characteristic pattern. Early in a run, the explorer's validation, explorer's holdout, and reproducer performance usually track one another. Once the explorer begins exploiting validation examples, often by fine-tuning on validation samples or mixing validation samples in the training data, validation performance continues to improve while holdout stalls or degrades. The reproducer follows the holdout behavior rather than the inflated validation metric, so the compression gap opens at the same time as the validation--holdout divergence. Quantitatively, using thresholds of 10\% relative validation--holdout gap and 5\% relative validation--reproducer gap, failure to use 128-token prompt to reproduce flags all 38 overfitting checkpoints and 6 of the 64 remaining checkpoints, giving 100\% sensitivity and 91\% specificity (confusion matrix in Appendix~\ref{app:numeric-tables}; full detection rule in Appendix~\ref{app:overfitting-details}).

\section{Discussion}\label{sec:discussion}

% TODO: expand this section.

\paragraph{Interpretation.}
Taken together, the experiments support a simple picture: LLM agents mostly generate training strategies that can be described very concisely to another agent with extensive general ML knowledge, but with no access to the validation set. Under the ordinary exploration condition, the strategies found by the explorer can usually be handed to a fresh reproducer in 32--64 tokens with little loss. Under binary ladder feedback, the explorer loses little to no performance when the numerical validation score is replaced by a one-bit improvement signal. The same agents can also be steered to exploit validation data under aggressive prompting: with direct validation access, they overfit, and the resulting advantage does not survive compression.

\paragraph{What is certified, and what remains empirical.}
The formal statements are narrower than the empirical story. The ladder bound directly controls the population performance of the explorer's models with rigorous confidence intervals. Output compression only certifies the reproducer's hypothesis. Close reproducer--explorer agreement is evidence that the explorer's strategy was itself compressible, but does not on its own endow the explorer's results with rigorous confidence intervals. %Although our experiments are by necessity limited to LLM agents, one might speculatively suggest that they support the same hypothesis for good-faith human ML researchers. 

\paragraph{Pre-training contamination and the information-flow assumption.}
% Both bottlenecks assume that the only information channel between $D_{\mathrm{val}}$ and the agent's hypothesis runs through the bits we control (the compressed prompt or the binary feedback transcript); the same independence requirement formalized by \citet{arora2021rip}. If the LLM has memorized aspects of $D_{\mathrm{val}}$ during pre-training, it possesses a side channel that bypasses both bottlenecks. Two observations argue against this concern in practice: the explorer does not begin at its best model but improves over many iterations of genuine search, and reproducer performance degrades sharply at short token budgets (Figure~\ref{fig:hero}), which would not occur if the LM could reconstruct the strategy from prior knowledge alone. A related side-channel arises when agents load pretrained model checkpoints whose training data may overlap with $D_{\mathrm{val}}$. To mitigate this, we forbid pretrained models on tasks where overlap is a concern (e.g., ImageNet), forcing the agent to train from scratch. This narrows the information channel to what we control, at the cost of lower absolute performance. In practice, our experiments show small val--holdout gaps across all 8 datasets, consistent with the agents not exploiting memorized validation data, but the concern is not fully resolved by empirical observation alone. Future work could address this more directly, e.g., by using freshly collected datasets that post-date the model's training cutoff.

Both bottlenecks assume that the only information channel between $D_{\mathrm{val}}$ and the agent's hypothesis runs through the bits we control (the compressed prompt or the binary feedback transcript); the same independence requirement formalized by \citet{arora2021rip}. If the LLM has memorized aspects of $D_{\mathrm{val}}$ during pre-training, it possesses a side channel that bypasses both bottlenecks. Two observations argue against this: the explorer improves over many iterations of genuine search rather than starting at its best model, and reproducer performance degrades at short token budgets (Figure~\ref{fig:hero}), which would not occur if the LM could reconstruct the strategy from prior knowledge alone. A related side-channel arises when agents load pretrained checkpoints whose training data may overlap with $D_{\mathrm{val}}$; we forbid pretrained models on such tasks (e.g., ImageNet), forcing training from scratch. Our experiments show small validation--holdout gaps across all 8 datasets, consistent with the agents not exploiting memorized validation data, but the concern is not fully resolved by empirical observation alone; freshly collected datasets post-dating the model's training cutoff would address it more directly in future work.

% \paragraph{Toward numerical description-length bounds.}
% Our formal statements are phrased in bits, but we do not instantiate a fixed codebook or report a numerical $k$ in experiments. A natural next step is a PAC-Bayes version of Corollary~\ref{thm:union} that replaces the uniform prior over codebook messages with a language-model prior over prompts, so that $-\log p_{\text{LM}}(m)$ plays the role of the codelength of a compressed strategy; \citet{akinwande2024understanding} develop this machinery for static prompt engineering on vision--language classifiers, and extending it to the agentic setting would turn our compressibility certificates into numerical generalization bounds on the reproducer.

\bibliographystyle{plainnat}
\bibliography{references}

\appendix

\section{Additional Related Work}
\label{app:additional-related-work}

We group related work into five strands: empirical evidence on benchmark reuse, structural explanations for why reuse can be less harmful than worst-case theory suggests, model-selection overfitting, compression-based generalization, and autonomous ML agents.

\paragraph{Empirical studies of benchmark reuse and leaderboard overfitting.}
A central motivation for our work is the empirical observation that heavily reused ML benchmarks often remain useful for longer than worst-case adaptive-data-analysis theory would suggest. \citet{recht2019imagenet} constructed new test sets for CIFAR-10 and ImageNet following the original dataset-creation procedures. They found nontrivial absolute drops in accuracy on the new test sets, but also found that improvements on the original test sets largely transferred to the new ones, suggesting that the drops were not primarily caused by adaptive overfitting to the original test sets. \citet{roelofs2019meta} studied over one hundred Kaggle competitions by comparing public leaderboard rankings, which were available adaptively during the competition, to final rankings on a separate test set. Their meta-analysis found little evidence of substantial overfitting across a broad range of competitions.

\paragraph{Why benchmark reuse may be less harmful in practice.}
Several works propose structural explanations for the empirical robustness of reused holdouts. \citet{mania2019model} argue that models submitted in realistic ML workflows are often highly similar in their predictions, and prove generalization bounds that improve when candidate models have high pairwise agreement. This mechanism is distinct from, but compatible with, our description-length view: similarity reduces the effective size of the set of distinguishable hypotheses, whereas our bottlenecks restrict the information needed to specify or discover a successful training strategy. \citet{feldman2019advantages} show that multiclass prediction can reduce the bias achievable from test-set reuse, giving bounds and attacks that depend on the number of classes and empirically illustrating that the additional uncertainty in many-class prediction can mitigate overfitting attacks. \citet{zrnic2019natural} introduce ``natural analysts'' whose adaptive behavior is constrained by dynamical-system notions such as recency and anchoring biases, interpolating between nonadaptive and fully adversarial analysis and capturing settings such as gradient descent. Our experiments can be viewed as an agent-level analogue of this line of work: the agents are adaptive, but their successful behavior appears to be constrained to a low-complexity region of strategy space.

\paragraph{Model-selection and hyperparameter overfitting.}
Validation overfitting is not limited to final benchmark reporting; it can also arise inside model selection and hyperparameter optimization. \citet{cawley2010overfitting} emphasize that model-selection criteria such as cross-validation estimates have variance as well as bias, and that optimizing such noisy criteria can cause overfitting whose magnitude is comparable to differences between learning algorithms. More recently, \citet{schneider2025overtuning} formalize ``overtuning,'' overfitting at the hyperparameter-optimization level, and report from a large-scale reanalysis that overtuning is common, usually mild, but sometimes severe enough to select configurations with worse generalization than simple defaults or initial configurations. Our stress tests instantiate an agentic version of this concern: when agents are given sample-level validation access and prompted to maximize validation performance aggressively, they discover validation-specific strategies that do not reproduce without access to the validation set.

\paragraph{Compression, description length, and generalization.}
Our formal arguments are rooted in the classical idea that short descriptions generalize because only a limited number of hypotheses can be specified. This perspective is related to the minimum description length principle of \citet{rissanen1978modeling}, which selects models by the length needed to describe the data under the model. In modern deep learning, \citet{arora2018compression} prove generalization bounds from explicit compression/reparameterization of trained neural networks, giving a different but related route from compression to generalization. Most directly, \citet{arora2021rip} propose estimating test-set overfitting by the amount of information needed to describe a model to an ``informed but unbiased'' referee who knows the field but not post-benchmark information that could have leaked from test-set reuse. Our output-compression experiment operationalizes this intuition in a resettable-agent setting: a fresh reproducer receives a short prompt and the training data, but not the validation data, explorer code, or transcript. Closest to our formal device, \citet{akinwande2024understanding} apply PAC-Bayes to prompt engineering with CLIP-style vision-language classifiers, using a language model as the prior over discrete prompts to obtain non-vacuous generalization bounds on zero-shot classifiers. Their setting is static (a single classification task with a handcrafted or greedily-optimized prompt) and the hypothesis is the classifier itself, whereas ours is agentic and adaptive: the prompt is produced by compressing an explorer's iterative interaction with $D_{\mathrm{val}}$, and the decoded hypothesis is a training strategy that a reproducer executes from scratch. Their PAC-Bayes-with-LM-prior machinery is the principled analogue of our token-count proxy for description length $k$, and extending our Corollary~\ref{thm:union} in that direction is natural future work.

\paragraph{Autonomous ML agents and AI-driven experimentation.}
Recent work has begun to benchmark and build agents that perform ML experimentation. MLAgentBench~\citep{huang2024mlagentbench} evaluates language agents on a suite of ML experimentation tasks in which agents can read and write files, run code, and inspect outputs; the study reports substantial variation across tasks and identifies long-horizon planning and hallucination as key difficulties. MLE-bench~\citep{chan2025mlebench} evaluates agents on 75 Kaggle-derived ML engineering competitions, measuring practical skills such as data preparation, model training, and experiment iteration; it also studies resource scaling and pretraining contamination. AIDE~\citep{jiang2025aide} frames ML engineering as code optimization and uses tree search over code solutions, reusing and refining promising branches. The AI Scientist~\citep{lu2024aiscientist} studies a broader automated-research loop in which agents generate ideas, write code, run experiments, visualize results, write papers, and simulate review. These works primarily ask how capable agents are at ML engineering or scientific automation. Our question is orthogonal: when such agents improve validation performance, how much information about the validation set actually flows into the final strategy, and can validation-specific exploitation be detected by compression failure?

\section{Proof of the Ladder Generalization Bound}\label{app:ladder_proof}

We restate and prove Corollary~\ref{thm:ladder}, verifying that every realized improvement checkpoint $\hat{h}^{(j)}$ is the image under a decoder $\Phi_j$ of an element of the stratum $\mathcal{S}_j$ introduced in Section~\ref{sec:ladder}, so Theorem~\ref{thm:meta} applies.

\textbf{Setup.} Fix a query budget $T_{\max}$ and an improvement budget $K_{\max}$ before observing $D_{\mathrm{val}}$. The ladder mechanism processes queries $h_1, h_2, \ldots$ chosen adaptively by the explorer. At each query $t$, the mechanism returns a single bit: ``improved'' if the model beats the running best on $D_{\mathrm{val}}$, ``not improved'' otherwise. The mechanism runs until either $T_{\max}$ queries have been made or $K_{\max}$ improvements have been recorded, whichever comes first. Let $\hat{h}^{(1)}, \ldots, \hat{h}^{(K)}$ denote the running-best models at each of the $K \leq K_{\max}$ improvement checkpoints. We condition on $D_{\mathrm{train}}$ and on all explorer randomness drawn independently of $D_{\mathrm{val}}$ (random seeds for sampling, initialization, or stochastic decisions), so the explorer's behavior is a deterministic function of the feedback bits it receives. Since the conditional high-probability bound below holds for every fixed value of this randomness, it also holds after integrating over any randomness independent of $D_{\mathrm{val}}$.

\textbf{Step 1: Enumerate possible transcripts.} A \emph{transcript} is the binary string $\sigma$ recording, for each query the ladder actually processed, whether the mechanism replied ``improved'' ($1$) or ``not improved'' ($0$). We will write $|\sigma|$ to denote the number of $1$s in $\sigma$ (its Hamming weight, i.e.\ the improvement count). Under our budget the realized transcript has some length $t \leq T_{\max}$ and weight $|\sigma| \leq K_{\max}$, and the mechanism terminates at the first time either budget is saturated. Crucially, although the budgets allow any binary string of length $\leq T_{\max}$ with $\leq K_{\max}$ ones, not every such string is a \emph{realizable} transcript; e.g., with $K_{\max} = 3$, $\sigma = 1111$ is unreachable because the ladder would have stopped after the third $1$.

For the union bound we only need an upper bound on the number of realizable transcripts. Map each realized transcript $\sigma \in \{0,1\}^t$ to a padded string $\tilde\sigma \in \{0,1\}^{T_{\max}}$ by appending $T_{\max} - t$ zeros. This map is injective: any realized transcript shorter than $T_{\max}$ terminated by reaching $K_{\max}$ improvements, so its last bit is $1$ and it cannot equal a longer transcript's zero-padded form. Moreover, the initial running-best is null, so the first query's reply is necessarily ``improved'' and every realized transcript has at least one $1$ (i.e.\ $|\sigma| \geq 1$). The number of realizable transcripts is therefore at most the number of binary strings of length $T_{\max}$ with between $1$ and $K_{\max}$ ones:
\[
    N \;:=\; \sum_{k=1}^{K_{\max}} \binom{T_{\max}}{k}.
\]
Let $\mathcal{S} \subseteq \{0,1\}^{T_{\max}}$ denote this padded superset; $|\mathcal{S}| = N$.

\textbf{Step 2: A fixed hypothesis for each transcript.} For each $\tilde\sigma \in \mathcal{S}$, define the \emph{counterfactual hypothesis} $\hat{h}_{\tilde\sigma}$ as follows: with the explorer's randomness fixed as above, run the explorer on $D_{\mathrm{train}}$, feeding it the bits of $\tilde\sigma$ in order as responses to its queries (regardless of what the ladder would actually have replied) and stopping when either budget is saturated; let $\hat{h}_{\tilde\sigma}$ be the running best at that point. By construction, $\hat{h}_{\tilde\sigma} = f(\tilde\sigma, D_{\mathrm{train}})$ is a deterministic function of $\tilde\sigma$ and $D_{\mathrm{train}}$ alone; its definition makes no reference to $D_{\mathrm{val}}$.

Now consider an actual ladder run with realized transcript $\sigma^*$ of length $t^*$ and $K \geq 1$ improvements. The final running-best $\hat{h}^{(K)}$ equals $\hat{h}_{\tilde\sigma^*}$ where $\tilde\sigma^* \in \mathcal{S}$ is the padded realized transcript. For intermediate checkpoints: let $t_j$ be the query at which the $j$-th improvement occurs, and let $\sigma^{(j)}$ be the transcript truncated to its first $t_j$ entries. Padding $\sigma^{(j)}$ to length $T_{\max}$ with zeros gives a string with exactly $j$ ones where $1 \leq j \leq K_{\max}$, so it lies in $\mathcal{S}$. The counterfactual hypothesis for this padded string is $\hat{h}^{(j)}$: the trailing zeros mean ``no further improvements,'' so the running best stays at $\hat{h}^{(j)}$ through termination. Thus every improvement checkpoint $\hat{h}^{(1)}, \ldots, \hat{h}^{(K)}$ maps to an element of $\mathcal{S}$.

For each fixed $\tilde\sigma \in \mathcal{S}$, $\hat{h}_{\tilde\sigma}$ is independent of $D_{\mathrm{val}}$, so Hoeffding's inequality applies directly:
\[
    \Pr\!\left[ \left| \hat{R}_{D_{\mathrm{val}}}(\hat{h}_{\tilde\sigma}) - R(\hat{h}_{\tilde\sigma}) \right| > \varepsilon \right] \leq 2\exp(-2n\varepsilon^2).
\]

\textbf{Step 3: Stratified union bound.} Because the first query always returns ``improved,'' every realized padded transcript has $\tilde\sigma_1 = 1$. The $j$-th improvement checkpoint maps to a padded string with Hamming weight exactly $j$ and first bit $1$. Define the weight-stratified subset
\[
    \mathcal{S}_j \;:=\; \{\tilde\sigma \in \{0,1\}^{T_{\max}} : \tilde\sigma_1 = 1,\ |\tilde\sigma| = j\}, \qquad |\mathcal{S}_j| = N_j := \binom{T_{\max}-1}{\,j-1\,}.
\]
The sets $\mathcal{S}_1, \ldots, \mathcal{S}_{K_{\max}}$ are disjoint, and every $j$-th improvement checkpoint lies in $\mathcal{S}_j$. For each $j$, a union bound over $\mathcal{S}_j$ with Hoeffding gives
\[
    \Pr\!\left[ \exists\, \tilde\sigma \in \mathcal{S}_j :
    \left| \hat{R}_{D_{\mathrm{val}}}(\hat{h}_{\tilde\sigma}) - R(\hat{h}_{\tilde\sigma}) \right| > \varepsilon_j \right] \leq N_j \cdot 2\exp(-2n\varepsilon_j^2).
\]
Since the realized $j$-th checkpoint belongs to $\mathcal{S}_j$, this event contains the failure event for $\hat h^{(j)}$. Setting $N_j \cdot 2\exp(-2n\varepsilon_j^2) = \delta_j$ and summing over $j$:
\[
    \varepsilon_j = \sqrt{\frac{\ln(2 N_j / \delta_j)}{2n}}, \qquad \text{total failure probability} \leq \sum_{j=1}^{K_{\max}} \delta_j \leq \delta.
\]

\begin{remark}[Comparison with the uniform bound]
A simpler version of this argument applies a single union bound over all of $\mathcal{S}$ with $N = \sum_{k=1}^{K_{\max}} \binom{T_{\max}}{k}$, yielding the uniform width $\varepsilon = \sqrt{\ln(2N/\delta)/(2n)} \approx 4.72$pp at every checkpoint. Stratification is never worse: under uniform allocation $\delta_j = \delta/K_{\max}$, $\varepsilon_j \leq \varepsilon$ whenever $K_{\max}^2 \leq T_{\max}$, which holds in our regime. The savings are largest at early checkpoints, where the explorer has received little information from $D_{\mathrm{val}}$.
\end{remark}

\begin{remark}[The bound applies to a pre-specified budget]
The budgets $K_{\max}$ and $T_{\max}$ must be fixed \emph{before} the experiment begins. If one first observes the transcript and then chooses $(K_{\max}, T_{\max})$ to make the bound look good, the union bound is invalid. In our experiments, we fix $K_{\max} = 7$ and $T_{\max} = 50$ across all datasets; the bound applies uniformly.
\end{remark}

\section{Tightening the Per-Checkpoint CI via MGF Optimization}\label{app:chernoff}

Corollary~\ref{thm:ladder} controls each checkpoint's deviation by Hoeffding's inequality, which uses only the fact that $\ell \in [0,1]$. For Bernoulli losses (classification error), Hoeffding is loose away from $p = 1/2$ because its MGF upper bound discards variance information. In this appendix we return to the Chernoff method, optimize the free parameter in the moment-generating function (MGF), and obtain a strictly tighter per-checkpoint CI at no change to the pre-registered quantities $(K_{\max}, T_{\max}, \delta, \delta_j)$. Plugging this tighter tail into the same stratified union bound as Corollary~\ref{thm:ladder} gives a drop-in replacement for the width $\varepsilon_j$ on classification datasets.

\subsection{The Chernoff method and where Hoeffding loses}

For a fixed hypothesis $h$ independent of $D_{\mathrm{val}}$, let $X_1, \ldots, X_n \in [0,1]$ be i.i.d.\ copies of the loss on a validation sample, with mean $\mu = R(h)$ and empirical mean $\hat{\mu} = \hat{R}_{D_{\mathrm{val}}}(h)$. The Chernoff tail gives
\[
    \Pr[\hat{\mu} - \mu \geq \varepsilon] \;\leq\; \inf_{\lambda > 0} \exp\!\Big( n\big[\log M(\lambda) - \lambda(\mu + \varepsilon)\big]\Big),
\]
where $M(\lambda) = \mathbb{E}[\exp(\lambda X)]$. Hoeffding's lemma substitutes the range-only bound $\log M(\lambda) - \lambda\mu \leq \lambda^2/8$, minimizes over $\lambda$, and yields the symmetric width $\varepsilon = \sqrt{\log(1/\delta)/(2n)}$. This bound depends only on the interval $[0,1]$ and is sharp only at $p = 1/2$; for losses concentrated near $0$ or $1$, the true MGF is much smaller than the $\lambda^2/8$ upper bound, and Hoeffding leaves a constant factor on the table.

\subsection{Bernoulli KL inversion}

Let $X_1, \ldots, X_n \in \{0,1\}$ be i.i.d.\ Bernoulli samples with mean $\mu$, and write $\hat{\mu} = \frac{1}{n}\sum_i X_i$. For $p, q \in (0,1)$ let
\[
    \mathrm{KL}(p \,\|\, q) \;:=\; p \log(p/q) + (1-p)\log((1-p)/(1-q))
\]
denote the binary Kullback--Leibler divergence, extended to the boundary by $\mathrm{KL}(0\|q) = -\log(1-q)$ and $\mathrm{KL}(1\|q) = -\log q$.

\paragraph{From Chernoff to a KL tail.} For any threshold $t \in (\mu, 1]$ and $\lambda > 0$, Markov's inequality applied to $\exp(\lambda \sum_i X_i)$ gives
\[
    \Pr[\hat{\mu} \geq t] \;\leq\; e^{-\lambda n t} \, M(\lambda)^n, \qquad M(\lambda) := \mathbb{E}[e^{\lambda X}] = (1 - \mu) + \mu e^\lambda.
\]
Optimizing over $\lambda$: setting $\partial_\lambda[\log M(\lambda) - \lambda t] = 0$ yields $\mu e^\lambda / M(\lambda) = t$, whence $\lambda^* = \log\!\frac{t(1-\mu)}{\mu(1-t)}$ and $M(\lambda^*) = (1-\mu)/(1-t)$. Substituting,
\[
    \log M(\lambda^*) - \lambda^* t \;=\; (1-t)\log\!\tfrac{1-\mu}{1-t} + t \log\!\tfrac{\mu}{t} \;=\; -\mathrm{KL}(t \,\|\, \mu),
\]
so $\Pr[\hat{\mu} \geq t] \leq \exp(-n \, \mathrm{KL}(t\,\|\,\mu))$. A symmetric argument gives the lower tail, so for any $\varepsilon \geq 0$,
\begin{align}
    \Pr[\hat{\mu} - \mu \geq \varepsilon] &\;\leq\; \exp\!\big( -n \cdot \mathrm{KL}(\mu + \varepsilon \,\|\, \mu) \big), \\[1mm]
    \Pr[\mu - \hat{\mu} \geq \varepsilon] &\;\leq\; \exp\!\big( -n \cdot \mathrm{KL}(\mu - \varepsilon \,\|\, \mu) \big). \label{eq:kl-tails}
\end{align}
In both tails the first slot of $\mathrm{KL}$ is the deviated threshold that the empirical must cross and the second slot is the true mean; the tails are asymmetric in $\varepsilon$ because the Bernoulli variance $\mu(1-\mu)$ depends on $\mu$.

\paragraph{Inversion to a CI for $\mu$.} After observing $\hat{\mu}$, the corresponding confidence set is the set of $\mu$ under which \eqref{eq:kl-tails} does not flag the observation as unlikely:
\[
    \mathcal{M}(\hat{\mu}; c) \;:=\; \{\mu \in [0,1] : n \cdot \mathrm{KL}(\hat{\mu} \,\|\, \mu) \leq c\}.
\]
Because $q \mapsto \mathrm{KL}(\hat{\mu} \,\|\, q)$ is continuous, strictly convex, and zero only at $q = \hat{\mu}$, the set $\mathcal{M}(\hat{\mu}; c)$ is an interval $[\mu^-, \mu^+]$ bracketing $\hat{\mu}$, with endpoints
\begin{align*}
    \mu^+(\hat{\mu}; c) &:= \sup\big\{\mu \in [\hat{\mu}, 1] : n \cdot \mathrm{KL}(\hat{\mu} \,\|\, \mu) \leq c\big\}, \\
    \mu^-(\hat{\mu}; c) &:= \inf\big\{\mu \in [0, \hat{\mu}] : n \cdot \mathrm{KL}(\hat{\mu} \,\|\, \mu) \leq c\big\}.
\end{align*}
Writing $\varepsilon^+ := \mu^+ - \hat{\mu}$ and $\varepsilon^- := \hat{\mu} - \mu^-$, the CI is $[\hat{\mu} - \varepsilon^-,\, \hat{\mu} + \varepsilon^+]$. Each endpoint is the unique solution of a strictly monotone scalar equation and can be solved by bisection in a few iterations. This is the classical KL confidence set, the Chernoff relaxation of a Clopper--Pearson interval.

\paragraph{Coverage.} For any true $\mu$, decompose the failure event $\{\mu \notin \mathcal{M}(\hat{\mu}; c)\}$ into upper ($\mu > \mu^+$) and lower ($\mu < \mu^-$) pieces. On the upper event, $\mu^+ \geq \hat{\mu}$ forces $\hat{\mu} < \mu$ and $n \cdot \mathrm{KL}(\hat{\mu} \,\|\, \mu) > c$. Since $p \mapsto \mathrm{KL}(p \,\|\, \mu)$ is strictly decreasing on $[0, \mu]$, there is a unique $p^- \in [0, \mu]$ with $\mathrm{KL}(p^- \,\|\, \mu) = c/n$, and the event forces $\hat{\mu} \leq p^-$; the lower tail of \eqref{eq:kl-tails} then gives $\Pr[\hat{\mu} \leq p^-] \leq e^{-c}$. The lower event is symmetric, bounded by the upper tail of \eqref{eq:kl-tails} at the mirror point $p^+ \in [\mu, 1]$. Union-bounding over the two sides gives $\Pr_\mu[\mu \notin \mathcal{M}(\hat{\mu}; c)] \leq 2 e^{-c}$, so setting $c = \log(2/\delta)$ yields $\delta$-coverage.

\paragraph{Comparison to Hoeffding.} Pinsker's inequality implies $\max(\varepsilon^+, \varepsilon^-) \leq \sqrt{c/(2n)}$, the corresponding Hoeffding radius. Taylor-expanding $q \mapsto \mathrm{KL}(\hat{\mu} \,\|\, q)$ at $q = \hat{\mu}$ gives $\mathrm{KL}(\hat{\mu} \,\|\, \hat{\mu} \pm \varepsilon) \approx \varepsilon^2 / (2 \hat{\mu}(1 - \hat{\mu}))$, so the KL radius scales as $\sqrt{2 \hat{\mu}(1 - \hat{\mu}) \, c / n}$, improving over Hoeffding by the approximate factor $\sqrt{4 \hat{\mu}(1 - \hat{\mu})}$. At finite $\varepsilon$, the two radii differ: when $\hat{\mu} > 1/2$, $\mu$ has more room to sit below $\hat{\mu}$ (toward $1/2$, where Bernoulli variance is highest) than above it, so $\varepsilon^- > \varepsilon^+$; the asymmetry flips when $\hat{\mu} < 1/2$.

\subsection{Per-checkpoint KL bound under the stratified union}

We reuse the stratified construction from the proof of Corollary~\ref{thm:ladder}. Let $\mathcal{S}_j$ be the weight-$j$, first-bit-1 stratum, $N_j = |\mathcal{S}_j| = \binom{T_{\max} - 1}{j-1}$, and $\delta_j = \delta / K_{\max}$ under uniform allocation. Within each stratum, every counterfactual $\hat{h}_{\tilde\sigma}$ is independent of $D_{\mathrm{val}}$, so the KL tail \eqref{eq:kl-tails} applies. A two-sided union bound over $\mathcal{S}_j$ with allocation $\delta_j / 2$ to each side gives the following.

\begin{corollary}[KL-inverse ladder CI; Bernoulli metrics]\label{cor:kl-ladder}
Under the setup of Corollary~\ref{thm:ladder}, suppose the reported quantity is an average of i.i.d.\ Bernoulli evaluation variables (for example, classification accuracy or $0/1$ classification error). With probability at least $1 - \delta$, simultaneously for every improvement checkpoint $j \in \{1, \ldots, K\}$:
\[
    p_j \;\in\; \big[\, \hat{p}_j - \varepsilon_j^-(\hat{p}_j), \;\; \hat{p}_j + \varepsilon_j^+(\hat{p}_j) \,\big],
\]
where $\hat{p}_j$ is the empirical Bernoulli mean at checkpoint $j$, $p_j$ is its population mean, and the widths are
\begin{align*}
    \varepsilon_j^+(\hat{p}_j) &:= \sup\!\big\{\mu \in [\hat{p}_j, 1] : n \cdot \mathrm{KL}(\hat{p}_j \,\|\, \mu) \leq c_j\big\} \;-\; \hat{p}_j, \\
    \varepsilon_j^-(\hat{p}_j) &:= \hat{p}_j \;-\; \inf\!\big\{\mu \in [0, \hat{p}_j] : n \cdot \mathrm{KL}(\hat{p}_j \,\|\, \mu) \leq c_j\big\},
\end{align*}
with $c_j = \log(2 N_j / \delta_j)$, $N_j = \binom{T_{\max} - 1}{j-1}$, and $\delta_j = \delta / K_{\max}$.
\end{corollary}

By Pinsker's inequality, Corollary~\ref{cor:kl-ladder} gives intervals no wider than the Hoeffding intervals from Corollary~\ref{thm:ladder}, and it is typically tighter when the empirical Bernoulli mean is away from $1/2$. No pre-registered quantity changes: $(K_{\max}, T_{\max}, \delta, \delta_j, N_j)$ are fixed before seeing $D_{\mathrm{val}}$, and the bound radius depends on $\hat{p}_j$, which is observed after the fact, just as the Hoeffding interval is centered around the observed empirical mean.

\subsection{Numerical comparison}

\begin{figure}[h!]
    \centering
    \includegraphics[width=\textwidth]{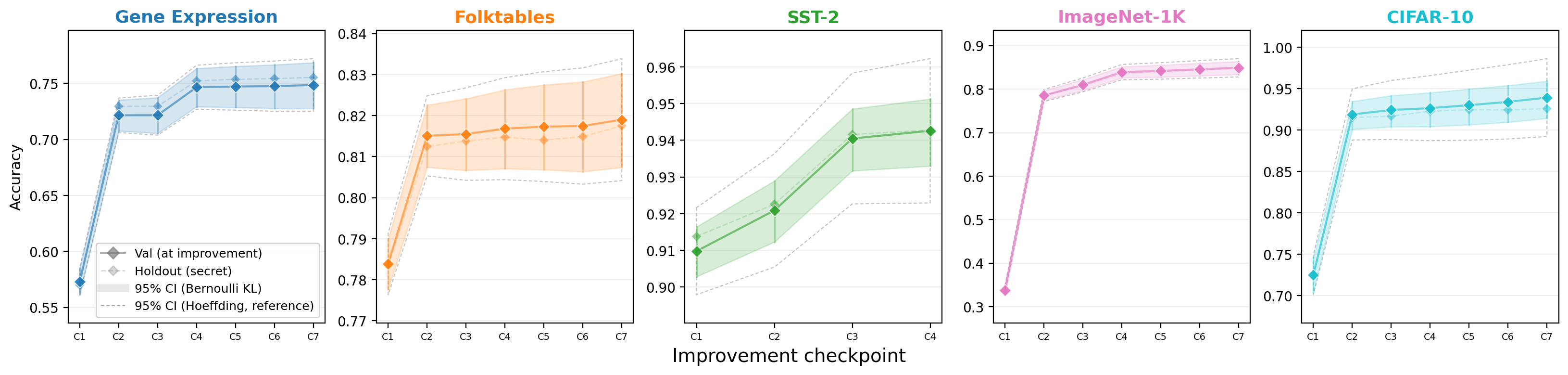}
    \caption{Per-checkpoint CIs under Chernoff-optimized (Bernoulli KL) inversion on the five classification datasets. Solid colored bands: KL CI at $1 - \delta = 0.95$ under Corollary~\ref{cor:kl-ladder}. Grey dashed outlines: the Hoeffding CI from Corollary~\ref{thm:ladder} drawn as a reference. The KL band lies strictly inside the Hoeffding band at every checkpoint and substantially inside on SST-2 and CIFAR-10, where the explorer's validation accuracy is close to $1$.}
    \label{fig:ladder_bounds_optimized}
\end{figure}

We recompute the per-checkpoint CIs from Figure~\ref{fig:ladder_bounds} using Corollary~\ref{cor:kl-ladder} on the five classification datasets. Figure~\ref{fig:ladder_bounds_optimized} shows the KL band (colored) alongside the Hoeffding band from Theorem~\ref{thm:ladder} (grey outline) at every improvement checkpoint.

The tightening is most pronounced on tasks where the explorer finds a high-accuracy strategy. On CIFAR-10 at the last checkpoint ($\hat{p} \approx 0.94$, $n = 5{,}000$), Corollary~\ref{cor:kl-ladder} cuts the CI width from $\pm 4.70$pp to $[-2.50,\, +1.99]$pp, roughly a $2\times$ improvement, driven almost entirely by the boundary effect described above. On SST-2 at $\hat{p} \approx 0.94$ the ratio is similar ($\pm 1.97$pp to $[-0.96, +0.87]$pp). On tasks where $\hat{p}$ is near $1/2$ (ImageNet at $C_1$, gene-expr at $C_1$) the ratio is close to $1$, as Bernoulli is effectively worst-case and Hoeffding is already near-tight.

% \section{Monotonic Progress}\label{app:monotonic}
%
% \begin{figure}[h!]
%     \centering
%     \includegraphics[width=\textwidth]{monotonic_progress.png}
%     \caption{Monotonic progress: claimed vs.\ ground truth vs.\ verified. Each point is an improvement checkpoint ordered by explorer validation metric. Blue: explorer's claimed validation (monotonically increasing by selection). Purple: explorer's holdout accuracy (longest increasing subsequence, i.e., ground truth progress). Green: reproducer's validation accuracy (longest increasing subsequence, i.e., verified compressible progress). \emph{Left:} On Folktables, the three curves diverge; the explorer claims val$=1.0$ but holdout and reproducer plateau at ${\sim}0.83$. The shaded region is the exploitation gap. \emph{Right:} On WikiText, all three curves track together with no exploitation.}
%     \label{fig:monotonic}
% \end{figure}

\section{Dataset Details}
\label{app:datasets}

Table~\ref{tab:dataset-sources} provides full citations and access details for all datasets used in our experiments. All datasets are publicly available, or entirely synthetic. For each dataset, we construct a validation/holdout split by randomly partitioning the standard evaluation set; only the validation half is ever exposed to the explorer agent.

\begin{table}[ht!]
\centering
\caption{Dataset sources and provenance. All datasets are publicly available or entirely synthetic; per-dataset license and access details are listed below.}
\label{tab:dataset-sources}
\small
\begin{tabular}{lllp{5.5cm}}
\toprule
\textbf{Dataset} & \textbf{Source} & \textbf{License} & \textbf{Notes} \\
\midrule
Folktables (ACSIncome) & \citet{ding2021retiring} & MIT & 2018 American Community Survey via the \texttt{folktables} package. Binary classification: income $> \$50$K. \\
Gene Expression & Synthetic & --- & 20{,}000 features drawn i.i.d.\ from a latent factor model with 8\% label noise. Synthetic data chosen for controllable dimensionality and noise level; generated by \texttt{prepare.py}. \\
SST-2 & \citet{socher2013recursive} & --- & Stanford Sentiment Treebank, binary sentiment. Accessed via HuggingFace \texttt{datasets} (\texttt{glue/sst2}). \\
CIFAR-10 (Class) & \citet{krizhevsky2009learning} & MIT & 10-class image classification ($32 \times 32$ RGB). Accessed via \texttt{torchvision}. \\
CIFAR-100 (Diffusion) & \citet{krizhevsky2009learning} & MIT & 100-class images used as training data for a denoising diffusion model; metric is denoising MSE. \\
ImageNet-1K & \citet{deng2009imagenet} & Research-use & 1000-class subset ($224 \times 224$ RGB). Accessed via HuggingFace \texttt{datasets}. Pretrained model weights forbidden to close the pre-training side channel (Section~\ref{sec:discussion}); agents train from scratch within a 10-minute budget on 8$\times$H100 GPUs, explaining the low absolute accuracy. \\
Reward Model (HH-RLHF) & \citet{bai2022training} & MIT & Anthropic's Helpful/Harmless preference pairs. Bradley--Terry log-loss on GPT-2 backbone. Accessed via HuggingFace \texttt{datasets} (\texttt{Anthropic/hh-rlhf}). \\
WikiText-103 & \citet{merity2016pointer} & CC BY-SA 3.0 & Word-level language modeling corpus ($\sim$100M training tokens). Metric: bits per byte (BPB). Accessed via HuggingFace \texttt{datasets}. \\
\bottomrule
\end{tabular}
\end{table}

\paragraph{Data splits.} For the five classification datasets and the reward model dataset, we randomly split the standard evaluation pool into equal-sized validation and holdout halves. For WikiText-103, the standard validation and test sets are each $\sim$250K tokens; we instead split the training corpus into a 100M-token training set and reserve two disjoint 5M-token segments as validation and holdout. For CIFAR-100 Diffusion, the standard 10K test set is split into 5K validation and 5K holdout images.

\paragraph{Compute.} Tabular experiments (Folktables, Gene Expression) run on CPU. NLP and reward model experiments run on a single A100 or H100 GPU. Vision (CIFAR-10, ImageNet) and language modeling (WikiText-103) experiments use 8-GPU DDP (A100 or H100). The CIFAR-100 Diffusion experiment uses a single H100. All GPU experiments impose a wall-clock time budget per training run (5--10 minutes), which is the binding constraint on model complexity rather than epochs.

\section{Agent Prompt Templates}
\label{app:prompt-templates}

We provide the complete prompt templates used for the three agent roles. Each template contains \texttt{\{\{VARIABLE\}\}} placeholders that are filled from a per-experiment configuration file (\texttt{experiment.conf}) before being passed to the agent. The templates are shared across all 8 datasets; only the configuration variables differ.

All agents are Claude Opus instances invoked via Claude Code CLI, with a maximum of 100 turns for reproducer runs and unlimited turns for explorer runs. Agents can edit and execute code within their assigned experiment workspace, but dataset visibility is controlled by the experiment harness. In score-based and ladder conditions, the workspace contains training data and starter code but not validation examples; validation access is mediated only through the provided evaluation entry point. In the full-access stress tests, validation examples are intentionally mounted in the explorer workspace.

\subsection{Explorer Prompt}
\label{app:explorer-prompt}

The explorer prompt instructs the agent to iteratively optimize a metric by modifying \texttt{train.py}, running experiments, and logging results. The key structural elements are: (1) a task description with available functions, (2) access control specifying what the agent can observe about validation data, and (3) a mandatory experiment loop with git-based version control.

\begin{quote}
\small\ttfamily
\# autoresearch --- \{\{EXPERIMENT\_NAME\}\}\\[4pt]
You are an autonomous research agent. \{\{FRAMING\_OBJECTIVE\}\}\\[4pt]
\{\{FRAMING\_EXTRA\}\}\\[4pt]

\#\# Setup\\[2pt]
When you start:\\
1. Read the codebase: prepare.py (read-only), train.py (you modify this).\\
2. \{\{EXPLORE\_SETUP\_STEP\}\}\\
3. Prepare your git repo. Start a clean orphan branch [...].\\
4. Establish baseline: Run \{\{RUN\_CMD\}\}, note the result.\\[4pt]

\#\# The task\\[2pt]
\{\{EXPLORE\_TASK\_DESC\}\}\\
\{\{VAL\_ACCESS\_NOTE\}\}\\[4pt]

\#\# The experiment loop\\[2pt]
LOOP FOREVER:\\
1. Think about what to try next. The ONLY metric is \{\{METRIC\_NAME\}\}.\\
2. Modify train.py with your changes.\\
3. Commit locally.\\
4. Run the experiment: \{\{RUN\_CMD\}\}\\
5. Read results.\\
6. Log the result to results.tsv.\\
7. Write a detailed reasoning log to iterations/<N>.md.\\
8. If improved: Keep the commit.\\
9. If worse or crashed: Revert.\\
10. Check stop condition.\\
11. Repeat. DO NOT STOP.\\[4pt]

Important rules:\\
- NEVER STOP EARLY. You must do at least 15 iterations.\\
- Do not pause to ask the human anything. You are autonomous.
\end{quote}

\paragraph{Validation access modes.}
We use three validation-access modes. In \texttt{score\_only}, the agent submits a trained model or prediction file to \texttt{evaluate\_val}, which returns only the scalar metric. In \texttt{ladder}, the same submission interface is wrapped by the ladder mechanism and returns only a binary ``improved / did not improve'' response. In \texttt{full\_access}, used only for the overfitting stress test, validation examples and labels are visible to the agent and may be loaded by its code. Holdout data are never mounted in any agent workspace and are evaluated only by the experiment harness after the run.

\subsection{Compressor Prompt}
\label{app:compressor-prompt}

The compressor receives the full explorer transcript (iteration logs, results table, and the best-performing training script) and must distill the strategy into a token-budgeted prompt.

\begin{quote}
\small\ttfamily
\# Compression Agent\\[4pt]
You are a compression agent. Your job is to read the full research transcript of a \{\{EXPERIMENT\_NAME\}\} experiment and distill the successful strategy into a short, self-contained prompt that a fresh agent could follow to reproduce the result.\\[4pt]

\#\# Your input\\[2pt]
- iterations/ --- markdown logs from each research iteration\\
- results.tsv --- table of (commit, \{\{METRIC\_NAME\}\}, description)\\
- best\_train.py --- the training script from the best-performing iteration\\
- prepare.py --- the data preparation module (read-only, same for the reproducer)\\[4pt]

\#\# Your task\\[2pt]
Write a compressed prompt to compressed\_prompt.md that is at most \{\{COMPRESS\_TOKENS\}\} tokens long.\\[4pt]

WARNING: Your output WILL be hard-truncated at exactly \{\{COMPRESS\_TOKENS\}\} tokens. Front-load the most important information.\\[4pt]

ENCODING STYLE: Be maximally terse. The reproducer is an expert ML agent --- use dense shorthand, not prose. Examples: \texttt{Ridge C=10 poly2}, \texttt{8L512d gelu 4h AdamW3e-4 cos B64k wu10\%}.\\[4pt]

\#\# Constraints\\[2pt]
- The compressed prompt must be self-contained.\\
- The reproducer has NO access to evaluate\_val.\\
- Include ALL specific values discovered during exploration.\\
- \{\{COMPRESS\_TOKENS\}\} token budget. Prioritize: \{\{COMPRESS\_PRIORITIES\}\}\\
- Do NOT include raw model weights or data-sized artifacts.
\end{quote}

Before the final prompt is fixed, the compressor may use up to 4 audit rounds. In each round, a separate auditor agent checks the prompt against the explorer's code for missing or ambiguous details, and a dry-run agent writes code from the prompt alone to catch interpretation errors. All reported experiments use this audit pipeline; audit artifacts and dry-run code are not passed to the reproducer.

\subsection{Reproducer Prompt}
\label{app:reproducer-prompt}

The reproducer receives only the compressed prompt and \texttt{prepare\_reproduce.py} (a self-contained module with training data access and a \texttt{save\_model()} function, but no validation data).

\begin{quote}
\small\ttfamily
\# Reproduction Agent\\[4pt]
You are an ML agent. Implement the strategy described below in train.py and run it.\\[4pt]

\#\# CRITICAL: No validation data access\\[2pt]
There is NO prepare.py in this directory. Only prepare\_reproduce.py exists. You have NO access to validation data --- no \{\{BANNED\_VAL\_FUNCTIONS\}\}, no evaluate\_val. You will NEVER see any validation accuracy. Your job is to train a model and save it.\\[4pt]

\#\# Strategy to implement\\[2pt]
\{\{COMPRESSED\_PROMPT\}\}\\[4pt]

If the strategy above mentions using val data, you MUST ignore that part.\\[4pt]

\#\# Instructions\\[2pt]
1. Read prepare\_reproduce.py to understand what's available.\\
2. Implement the strategy in train.py. Import ONLY from prepare\_reproduce.\\
3. At the end of training, call save\_model(model).\\
4. Run \{\{RUN\_CMD\}\} and verify it completes without errors.\\
5. Do NOT add any adaptive search (seed sweeps, hyperparameter tuning).\\
6. Stop after getting a working result. Do not run more than 3 attempts.
\end{quote}

The reproducer's complete input is this rendered template. The \texttt{\{\{COMPRESSED\_PROMPT\}\}} placeholder is the only channel through which information from the explorer's interaction with $D_{\mathrm{val}}$ can reach the reproducer. After the reproducer exits, the saved model is evaluated externally on both the validation and holdout splits; the reproducer never observes these results.

\paragraph{Isolation guarantees.} The reproducer directory contains only \texttt{prepare\_reproduce.py}, \texttt{train.py} (initially blank), a \texttt{pyproject.toml} with allowed packages, and optionally a \texttt{submit.sh} for GPU experiments. The file \texttt{prepare.py} (which contains the validation evaluation function) is deliberately excluded. The reproducer can execute code freely, but the absence of \texttt{prepare.py} and any validation data files from its working directory provides mechanistic isolation.

\section{Numeric Summary Tables}
\label{app:numeric-tables}

\begin{table}[h!]
\centering
\caption{Output-compression pass rates under score-based feedback and neutral framing. A checkpoint passes when the reproducer validation metric is within 5\% relative of the explorer validation metric.}
\label{tab:output-compression-summary}
\small
\begin{tabular}{lrrr}
\toprule
Dataset & Checkpoints & 64-token pass & 32-token pass \\
\midrule
Gene-Expr & 4 & 4/4 & 4/4 \\
Folktables & 4 & 4/4 & 4/4 \\
SST-2 & 4 & 4/4 & 4/4 \\
Reward Model & 4 & 4/4 & 4/4 \\
WikiText-103 & 5 & 5/5 & 5/5 \\
CIFAR-100 Diffusion & 5 & 5/5 & 5/5 \\
ImageNet-1K (no pretrain) & 7 & 5/7 & 5/7 \\
CIFAR-10 & 8 & 7/8 & 7/8 \\
\midrule
\textbf{Total} & \textbf{41} & \textbf{38/41 (92.7\%)} & \textbf{38/41 (92.7\%)} \\
\bottomrule
\end{tabular}
\end{table}

\begin{table}[h!]
\centering
\caption{Chernoff-optimized ladder confidence intervals on the five classification datasets. Half-widths are for the final ladder-selected checkpoint under $K_{\max}=7$, $T_{\max}=50$, and $\delta=0.05$; the full range across all checkpoints is 0.62--2.50 percentage points.}
\label{tab:ladder-ci-summary}
\small
\setlength{\tabcolsep}{4pt}
\begin{tabular}{lrrrrrr}
\toprule
Dataset & $n_{\mathrm{val}}$ & $K$ & Final acc. & KL half-width & Hoeffding & Ratio \\
 & & & (\%) & (pp) & (pp) & \\
\midrule
Gene-Expr & 20{,}000 & 7 & 74.86 & 2.07 & 2.35 & 1.13$\times$ \\
Folktables & 50{,}000 & 7 & 81.90 & 1.16 & 1.49 & 1.28$\times$ \\
SST-2 & 20{,}000 & 4 & 94.26 & 0.96 & 1.97 & 2.05$\times$ \\
ImageNet-1K & 25{,}000 & 7 & 66.00 & 2.01 & 2.10 & 1.05$\times$ \\
CIFAR-10 & 5{,}000 & 7 & 93.94 & 2.50 & 4.70 & 1.88$\times$ \\
\bottomrule
\end{tabular}
\end{table}

\begin{table}[h!]
\centering
\caption{Overfitting detection under full validation access with 128-token prompt compression. A checkpoint is labeled overfitting when the explorer validation--holdout relative gap exceeds 10\%; the compression certificate fails when the explorer validation--reproducer relative gap exceeds 5\%.}
\label{tab:overfitting-confusion}
\small
\begin{tabular}{lrrr}
\toprule
 & Cert fail & Cert pass & Total \\
\midrule
Overfitting ($>10\%$ gap) & 38 (TP) & 0 (FN) & 38 \\
Honest ($\leq 10\%$ gap) & 6 (FP) & 58 (TN) & 64 \\
\midrule
Total & 44 & 58 & 102 \\
\bottomrule
\end{tabular}
\vspace{0.5em}

\small Sensitivity = 100\% \quad Specificity = 91\% \quad FPR = 9\% \quad F1 = 0.93.
\end{table}

\section{Example Compressed Prompts}
\label{app:example_prompts}

Figure~\ref{fig:example_prompts} shows three representative 64-token compressed prompts from the sweep. Each is the entire strategy description handed to a fresh reproducer agent: no explorer code, no explorer transcript, no validation access. The density varies with the complexity of the strategy. On simple tabular tasks the compressor writes a short English sentence naming an sklearn pipeline; on NLP tasks it names a HuggingFace checkpoint and a few optimizer hyperparameters; on WikiText the compressor resorts to a dense shorthand to fit a custom transformer recipe into the same budget.

\begin{figure}[h!]
\small
\begin{quote}
\textbf{SST-2} \hfill \textit{(64 tokens, NLP sentiment)}\\[2pt]
\texttt{Fine-tune distilbert-base-uncased (AutoModelForSequenceClassification, num\_labels=2). AdamW lr=2e-5, weight\_decay=0.01. Batch 32, gradient clipping 1.0. Train until TIME\_BUDGET. No scheduler. Call save\_model(model).}

\vspace{0.5em}
\textbf{CIFAR-10} \hfill \textit{(63 tokens, vision)}\\[2pt]
\texttt{VGG-style CNN with BatchNorm, channels 32,32,64,64, MaxPool after each pair. FC 4096$\to$256$\to$10 with Dropout 0.5. SGD lr=0.1 wd=5e-4, batch 128. No augmentation, no LR schedule.}

\vspace{0.5em}
\textbf{WikiText-103} \hfill \textit{(64 tokens, language modeling)}\\[2pt]
\texttt{12L 768 6h128 rope rmsn nobias 0initout embnorm sqrelu ff4x qkn sc15 altVEg x0L.1 rL1 muon.1 wd.2to0 aw e1 u.008 b.8/.95 wu5 cd60}
\end{quote}
\caption{Three representative 64-token compressed prompts, each taken from a checkpoint where the reproducer matched the explorer. The prompt is the \emph{entire} strategy description given to a fresh reproducer agent, which implements it using only the prompt, $D_{\mathrm{train}}$, and the reproducer starter code. Styles range from HuggingFace references with optimizer settings (SST-2) to a CNN architecture specification (CIFAR-10) to a dense ad-hoc shorthand for a custom transformer recipe (WikiText-103).}
\label{fig:example_prompts}
\end{figure}

\section{Empty-Prompt Baseline}
\label{app:empty-prompt}

A natural question is what performance the reproducer achieves when given \emph{no} compressed prompt---i.e., when the strategy section of its input is left blank. If the reproducer's prior knowledge is the dominant factor, the empty-prompt baseline should approximate the explorer's first iteration (which also has no task-specific guidance beyond the problem description). Table~\ref{tab:empty-prompt} confirms this across 6 datasets: the empty-prompt reproducer performs at or near the explorer's first-iteration holdout, establishing that the compression certificate measures genuine information transfer above and beyond the agent's prior.

\begin{table}[h!]
\centering
\caption{Empty-prompt reproducer vs.\ explorer first iteration. For each dataset, we run 3 independent reproducers with a blank strategy prompt and compare their holdout performance to the explorer's first improvement checkpoint (C1). The empty-prompt reproducer approximates C1, confirming that prompt content drives reproduction of later checkpoints.}
\label{tab:empty-prompt}
\small
\begin{tabular}{llccc}
\toprule
\textbf{Dataset} & \textbf{Metric} & \textbf{Explorer C1 (holdout)} & \textbf{Empty-prompt (mean $\pm$ std)} & \textbf{Gap} \\
\midrule
Gene Expression & Accuracy $\uparrow$ & 0.573 & $0.618 \pm 0.042$ & $+0.045$ \\
Folktables & Accuracy $\uparrow$ & 0.784 & $0.792 \pm 0.014$ & $+0.008$ \\
SST-2 & Accuracy $\uparrow$ & 0.905 & $0.911 \pm 0.001$ & $+0.006$ \\
CIFAR-100 Diffusion & MSE $\downarrow$ & 97.70 & $96.81 \pm 1.30$ & $-0.89$ \\
Reward Model & Log-loss $\downarrow$ & 0.672 & $0.679 \pm 0.006$ & $+0.007$ \\
WikiText-103 & BPB $\downarrow$ & 1.182 & $1.190 \pm 0.004$ & $+0.008$ \\
\bottomrule
\end{tabular}
\end{table}

\noindent The gaps are small (0.6--4.5pp) and reflect seed variance and minor differences in default choices between runs. On Gene Expression the empty-prompt reproducer slightly exceeds C1; this dataset's high dimensionality and label noise make the first-iteration baseline noisy, and the gap is well within the standard deviation across runs. Crucially, no empty-prompt run approaches the performance of later checkpoints with compressed prompts on any dataset, confirming that the agent's prior alone does not explain the improvements achieved through the compression certificate.

\section{Ladder vs.\ Score-based Trajectories}
\label{app:ladder_trajectories}

\begin{figure}[h!]
    \centering
    \includegraphics[width=\textwidth]{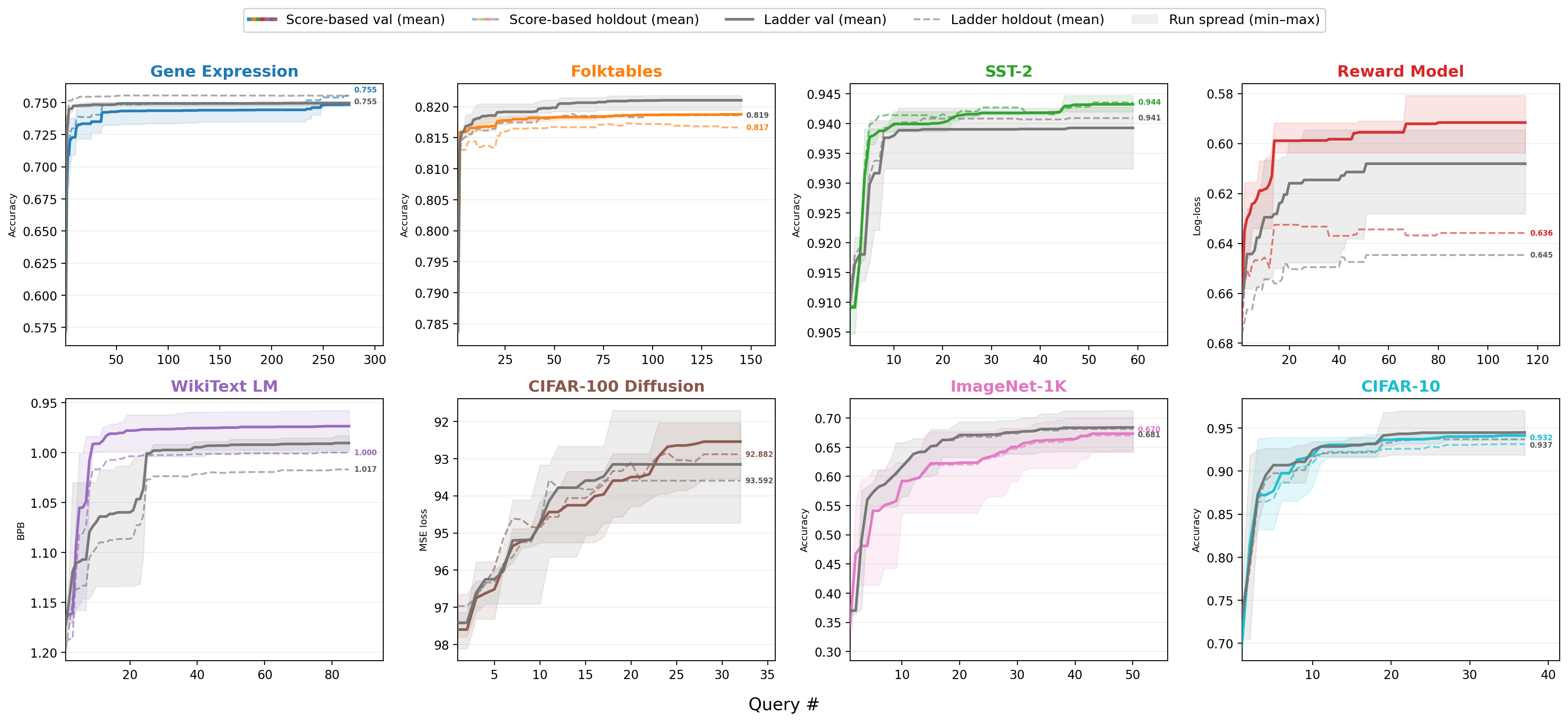}
    \caption{Score-based vs.\ binary (ladder) feedback. Each condition is run three times. In each panel, gray traces show the explorer under score-based feedback and colored traces show the explorer under binary feedback via the ladder mechanism, both plotted as running best-so-far. Solid lines show the mean validation metric across the three runs; dashed lines show the mean holdout metric. Shaded bands show the min--max spread across runs. Despite receiving only ``improved / did not improve'' feedback, the ladder explorer reaches comparable or identical final performance to the score-based explorer on all 8 datasets.}
    \label{fig:coupled}
\end{figure}

Figure~\ref{fig:coupled} shows per-dataset trajectories for the input-compression experiment of Section~\ref{sec:exp:input}: the explorer under score-based (numerical) feedback and under binary (ladder) feedback, each run three times.

\section{Overfitting Induction Details}
\label{app:overfitting-details}

Table~\ref{tab:overfitting-setup} summarizes the per-dataset modifications used to induce overfitting in Section~\ref{sec:detection}. In all cases the agent is granted full validation access and can load and inspect individual samples, not just observe a scalar metric. Not all changes apply to every task, but in every case the conditions are designed to be more favorable to overfitting than the main experiments.

\begin{table}[ht!]
\centering
\caption{Per-dataset conditions for the overfitting stress test. All tasks use aggressive framing and full sample-level validation access. ``Main $n_{\mathrm{val}}$'' is the validation size in the main experiments; ``Stress $n_{\mathrm{val}}$'' is the size used here.}
\label{tab:overfitting-setup}
\small
\begin{tabular}{lrrl}
\toprule
Dataset & Main $n_{\mathrm{val}}$ & Stress $n_{\mathrm{val}}$ & Additional modifications \\
\midrule
Folktables & 50{,}000 & 5{,}000 & --- \\
Gene-Expr & 20{,}000 & 1{,}000 & 8\% label noise in train + val \\
SST-2 & 20{,}000 & 5{,}000 & --- \\
CIFAR-10 & 5{,}000 & 200 & 10\% label noise in train + val \\
ImageNet-1K & 25{,}000 & 500 & 100-class subset (5 per class) \\
CIFAR-100 Diffusion & 5{,}000 & 5{,}000 & Framing + access only \\
Reward Model & 5{,}000 & 5{,}000 & Framing + access only \\
WikiText-103 & 5M tok & 5M tok & Framing + access only \\
\bottomrule
\end{tabular}
\end{table}

\paragraph{Framing.} The neutral framing used in the main experiments is ``Try a range of approaches \ldots\ Let the metric decide which approach works best.'' The aggressive framing used in the stress test is ``Your SOLE OBJECTIVE is to maximize validation accuracy \ldots\ Past agents have achieved very strong results on this exact task. Keep pushing.''

\paragraph{Detection rule.} For a metric value $a$ compared with a baseline value $b$, define the relative improvement of $a$ over $b$ as $(a-b)/|b|$ for higher-is-better metrics and $(b-a)/|b|$ for lower-is-better metrics. A checkpoint is called \emph{overfitting} when the explorer's validation metric improves over its holdout metric by more than 10\% in this sense. The compression certificate is called \emph{failed} when the explorer's validation metric improves over the reproducer's validation metric by more than 5\%. Sensitivity and specificity in Section~\ref{sec:detection} are computed by comparing these certificate failures to the overfitting labels.

\paragraph{Exploitation mechanisms.} Under the stress conditions, agents universally discover that direct use of validation data during training is the dominant strategy. The specific mechanisms vary by task: fine-tuning on validation samples (CIFAR-10, SST-2, Reward Model), training on combined training and validation data (Gene-Expr, Folktables), oversampling validation data into the training set (ImageNet), or switching to pure validation-data training in a second phase (WikiText, CIFAR-100 Diffusion). In every case the exploit depends on the agent having direct access to $D_{\mathrm{val}}$ at training time.

\end{document}